\documentclass[runningheads]{llncs}
\usepackage{graphicx}
\usepackage{comment}
\usepackage{graphicx}
\usepackage{subcaption}
\usepackage{hyperref}
\usepackage{amsmath,amssymb} 
\usepackage{color}
\usepackage[export]{adjustbox}
\usepackage{xspace}
\usepackage{array,multirow}
\usepackage[inline]{enumitem}

\makeatletter
\DeclareRobustCommand\onedot{\futurelet\@let@token\@onedot}

\def\@onedot{\ifx\@let@token.\else.\null\fi\xspace}

\def\eg{\emph{e.g}\onedot} 
\def\ie{\emph{i.e}\onedot}

\def\etal{\emph{et al}\onedot}

\makeatother

\newcommand\blfootnote[1]{%
  \begingroup
  \renewcommand\thefootnote{}\footnote{#1}%
  \addtocounter{footnote}{-1}%
  \endgroup
}
\begin{document}
\pagestyle{headings}
\mainmatter
\def\ECCVSubNumber{1694}  
\title{Learning to Scale Multilingual Representations for Vision-Language Tasks}

\titlerunning{SMALR}
%
\author{Andrea Burns\inst{1} \and
Donghyun Kim\inst{1} \and
Derry Wijaya\inst{1} \and
Kate Saenko\inst{1,2} \and
Bryan A. Plummer\inst{1}}
\authorrunning{A. Burns et al.}
%
\institute{Boston University, Boston MA 02215, USA \and MIT-IBM Watson AI Lab, Cambridge MA 02142, USA\\
\email{\{aburns4,donhk,wijaya,saenko,bplum\}@bu.edu}}
\maketitle
\begin{abstract}
Current multilingual vision-language models either require a large number of additional parameters for each supported language, or suffer performance degradation as languages are added. In this paper, we propose a Scalable Multilingual Aligned Language Representation (SMALR) that supports many languages with few model parameters without sacrificing downstream task performance. SMALR learns a fixed size language-agnostic representation for most words in a multilingual vocabulary, keeping language-specific features for just a few. We use a masked cross-language modeling loss to align features with context from other languages. Additionally, we propose a cross-lingual consistency module that ensures predictions made for a query and its machine translation are comparable. The effectiveness of SMALR is demonstrated with ten diverse languages, over twice the number supported in vision-language tasks to date. We evaluate on multilingual image-sentence retrieval and outperform prior work by 3-4\% with less than 1/5th the training parameters compared to other word embedding methods.

\end{abstract}

\blfootnote{Project page: \url{http://ai.bu.edu/smalr}} 
\section{Introduction}
\begin{figure*}[t]
\centering
    \begin{subfigure}[t]{0.55\textwidth}
        \centering
        \includegraphics[width=7cm]{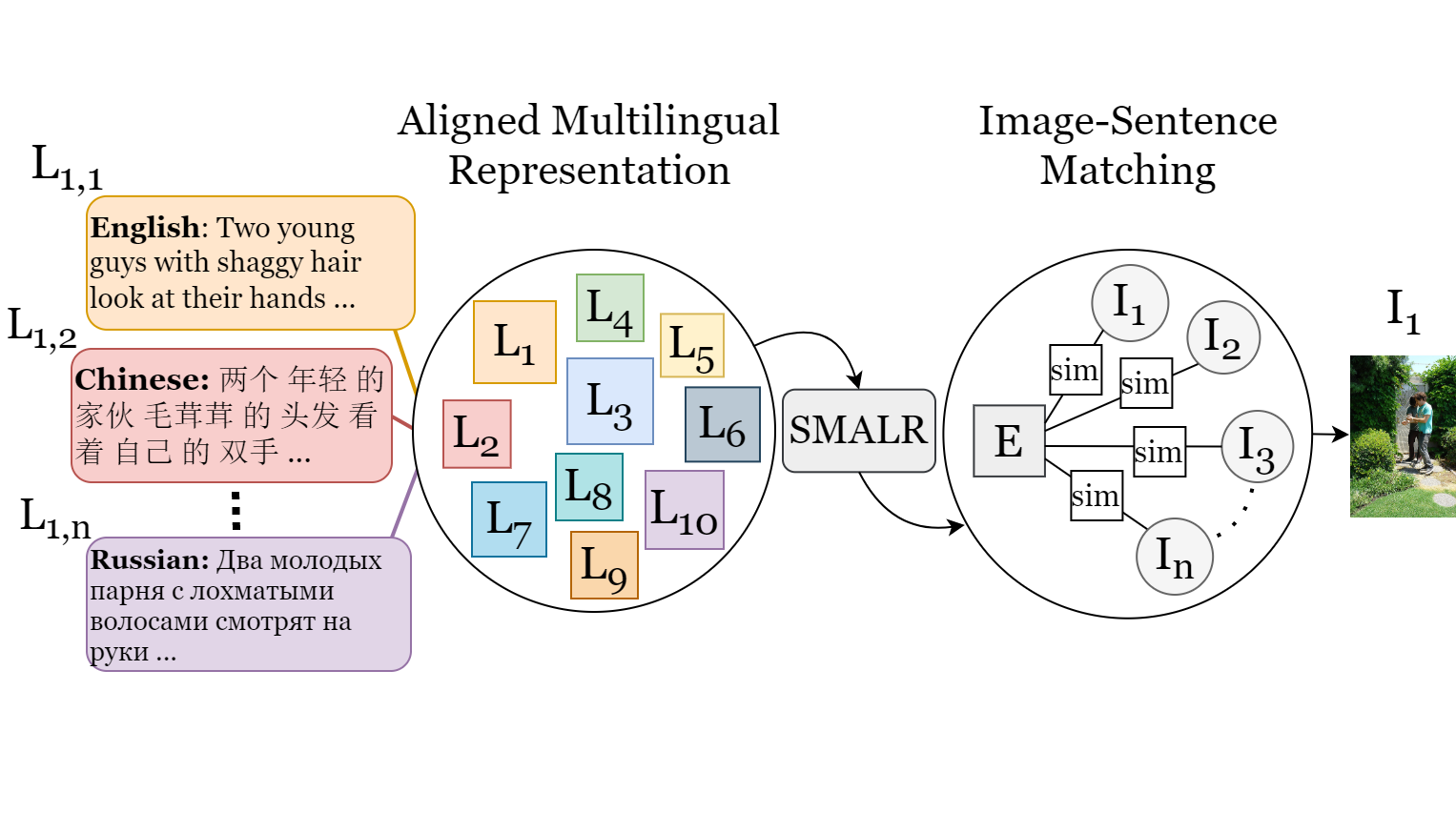}
        \caption{Multilingual image-sentence retrieval}
    \end{subfigure}%
    ~ 
    \begin{subfigure}[t]{0.45\textwidth}
        \centering
       \includegraphics[width=4.75cm]{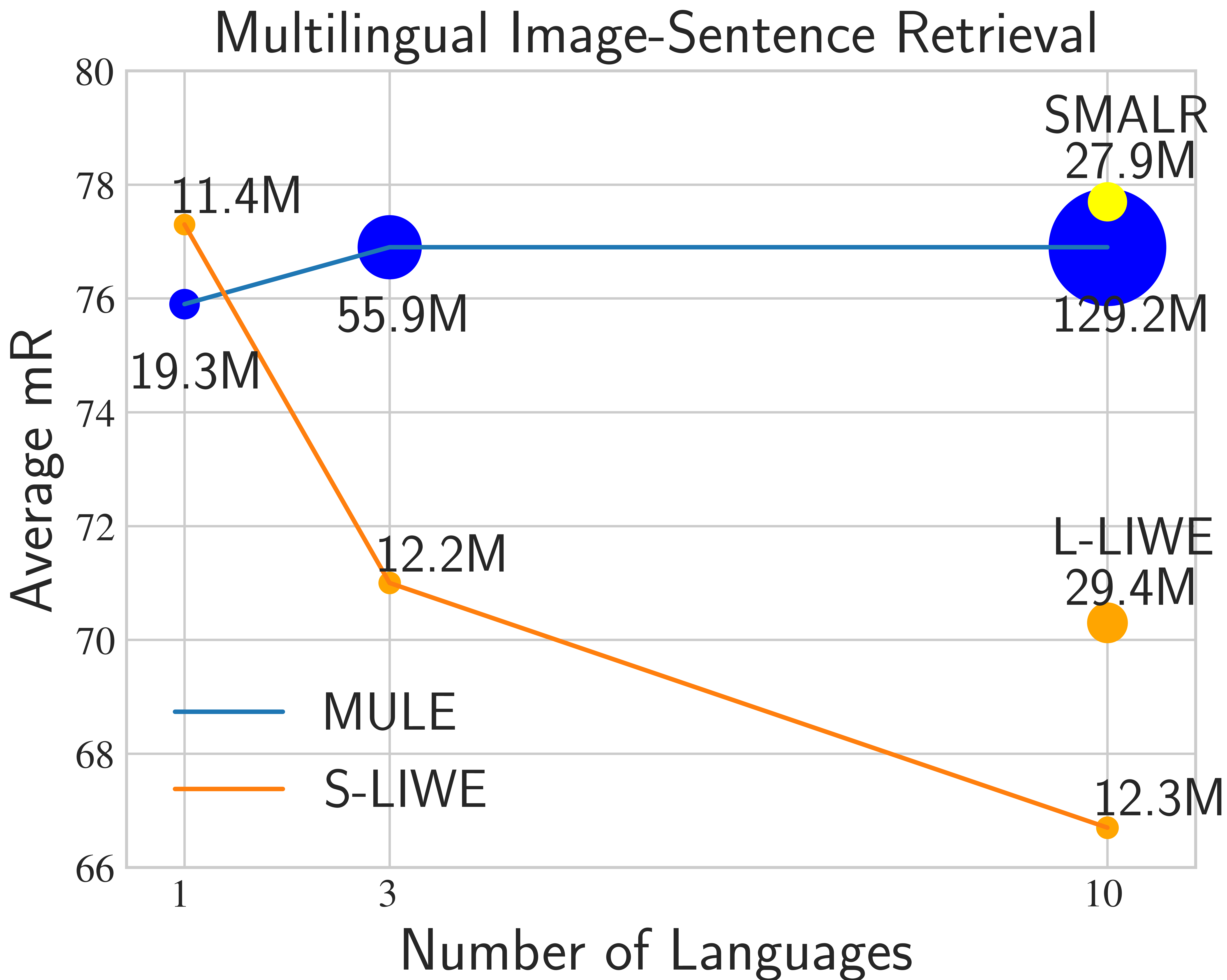}
        \caption{MSCOCO multilingual retrieval}
    \end{subfigure}
    
    \caption{(a) presents multilingual bidirectional retrieval. We embed sentences in ten languages with SMALR, which is used to compute the highest scoring image. (b) shows the effect of the number of training languages on performance for prior work MULE~\cite{kimMULEAAAI2020} and LIWE~\cite{Wehrmann_2019_ICCV}. LIWE is the original model, hereafter referred to as S-LIWE. The plot contains two points: L-LIWE,~\cite{Wehrmann_2019_ICCV} trained with a larger embedding (120-D vs. 24-D) for fair comparison, in orange, and SMALR, in yellow. The points are scaled to the number of parameters, P; specifically, their area is $(\frac{P}{10^6})^\frac{3}{2}$. SMALR is able to outperform all prior work with few parameters}
    \label{fig:motivation}
\end{figure*}

Learning a good language representation is a fundamental component of addressing a vision-language task, such as phrase grounding~\cite{kazemzadeh-EtAl:2014:EMNLP2014,plummer2015flickr30k} or visual question answering~\cite{AntolICCV2015,balanced_vqa_v2}. Many recent methods have demonstrated that learning
text representations aligned to images can boost performance across many vision-language tasks over traditional text-only trained representations~\cite{burnsLanguage2019,guptaEmbeddings2019,lu2019vilbert,su2019vlbert,tan2019lxmert}.  This is often accomplished by using auxiliary vision-language tasks when learning the language representation (such as image-sentence retrieval, as shown in Figure~\ref{fig:motivation}(a)).  However, these methods often only support a single language.  Although some work has addressed a multilingual scenario (\eg,~\cite{gellaEMNLP2017,kimMULEAAAI2020,Wehrmann_2019_ICCV}), these methods do not scale well to support many languages in terms of memory or performance (see Figure~\ref{fig:motivation}(b)).  As the number of languages grows, methods like LIWE~\cite{Wehrmann_2019_ICCV} that use character-based recognition systems can save memory but suffer from performance degradation. In contrast, methods that learn to align word embeddings across languages can maintain (or even improve) performance as languages are added (\eg,~\cite{gellaEMNLP2017,kimMULEAAAI2020}), but require additional parameters for the word embeddings that represent each new language's vocabulary.

This becomes a challenge when scaling to support many languages, as an increasing majority of trainable parameters are required for representing each language (\eg $\sim$93\% of parameters of~\cite{kimMULEAAAI2020} with ten languages). While pretrained word embeddings could be used without fine-tuning, \eg Multilingual BERT~\cite{bert} or MUSE~\cite{conneau2017word}, this comes at a significant cost in downstream task performance~\cite{burnsLanguage2019,kimMULEAAAI2020}.

To address this trade-off between multilingual capacity and performance, we propose a \textit{Scalable Multilingual Aligned Language Representation (SMALR)} model, which we demonstrate achieves strong task performance while also being highly compact compared to state-of-the-art word embedding methods~\cite{bert,klein2014fisher,li2019visualbert}. As seen in Figure~\ref{fig:motivation}, LIWE drops over 10\% in performance going from supporting one to ten languages. MULE slightly increases performance with more languages, but requires 6x more parameters compared to its single language model. Our approach, SMALR, outperforms both with only 1/5th the parameters of MULE.
We learn to efficiently represent each language by separating our language embedding into language-specific and language-agnostic token representations. As language follows a long-tailed distribution, only a few words occur often, with large portions of tokens occurring very rarely.  For example, in the MSCOCO dataset~\cite{lin2014microsoft} there are 25,126 unique tokens, but 61\% of them occur less than 4 times. This suggests that having unique representations for every token in the vocabulary is unnecessary, as only a subset would affect downstream task performance significantly.  Thus, we use a Hybrid Embedding Model (HEM) that contains language-specific embeddings for the common tokens, thereby providing a good representation for each language, and a compact language-agnostic representation for rare and uncommon words.  This results in a model that needs far fewer unique embeddings than prior work without sacrificing performance. 

\begin{figure}[t]
\centering
\includegraphics[scale=0.14]{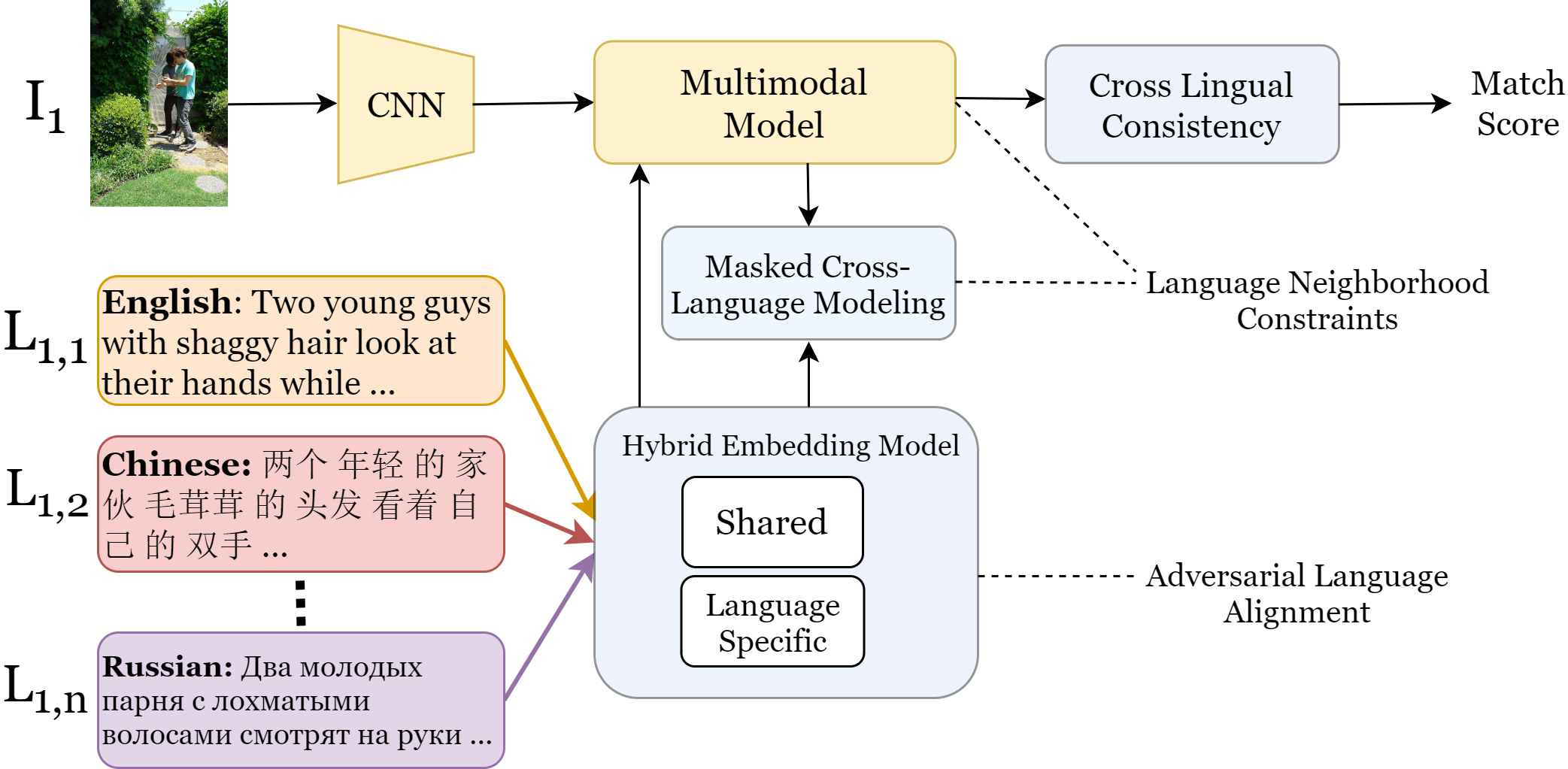}
\caption{The contributions of SMALR are in blue: a Hybrid Embedding Model (HEM), a Masked Cross-Language Model (MCLM), and a Cross-Lingual Consistency stage (CLC). HEM embeds input sentences as a mixture of language-specific and language-agnostic representations using a hard attention mechanism. The MCLM component provides an additional loss to enforce language alignment, while also augmenting the original dataset with masked sentences
}\label{fig:model}
\end{figure}

We learn how to assign tokens to the language-agnostic representation in a pretraining step, which uses monolingual FastText embeddings~\cite{bojanowski2017enriching} to map similar words to the same token, \eg~mapping ``double-decker" in English and ``imp\'eriale" in French to the same shared token. 
Once we obtain our language embeddings, our goal is to align them so that semantically similar words, even those from other languages, are embedded nearby.  To accomplish this, we use a multilingual masked language model, where we randomly mask words and then predict them based on context.  Unlike similar masking approaches used to train models such as BERT~\cite{bert}, we mask words of sentences from any two languages, say German and Chinese, which are semantically similar sentences referring to the same image, and use the context from each to predict both masked tokens.
To further encourage cross-language alignment, we also use an adversarial language classifier and neighborhood constraints that have been used in prior work~\cite{kimMULEAAAI2020}. These universal language embeddings are provided as input to a multimodal model that learns to relate them to images. Finally, we use a cross-lingual consistency module that uses machine translations to reason about the image-sentence similarity across multiple languages, which we show significantly boosts performance.  Figure~\ref{fig:model} contains an overview of our model. 

We use bidirectional image-sentence retrieval as the primary evaluation of our multilingual language representation.  In this task, the goal is to retrieve a relevant sentence from a database given an image or to retrieve a relevant image from a database given a sentence.  We augment current multilingual datasets Multi30K~\cite{barrault2018findings,elliott2017findings,elliott2016multi30k,young2014image} and MSCOCO~\cite{li2019coco,lin2014microsoft,miyazaki2016cross} using machine translations so that every image has at least five sentences across ten diverse languages: English (En), German (De), French (Fr), Czech (Cs), Chinese (Cn), Japanese (Ja), Arabic (Ar), Afrikaans (Af), Korean (Ko), and Russian (Ru).  See the supplementary for details on our data augmentation procedure. This constitutes the highest number of languages used in multilingual learning for vision-language tasks to date, supporting more than double the number of visually-semantically aligned languages compared to prior work~\cite{bilingual2,conneau2017word,gellaEMNLP2017,kimMULEAAAI2020,bilingual1,Wehrmann_2019_ICCV}.

We list the contributions of our work below:
\begin{itemize}
    \item SMALR, a scalable multilingual model for training visually-semantically aligned word embeddings that outperforms the state-of-the-art on multilingual image-sentence retrieval while also requiring few model parameters.
    \item A comparison to four types of vocabulary reduction methods that serve as baselines to complement our evaluation against prior work. 
    \item A Masked Cross-Language Modeling (MCLM) procedure that further aligns the multilingual embedding, stabilizing variance in performance over all languages, and serves as an additional data augmentation technique.
    \item A Cross-Lingual Consistency (CLC) module, the first of its kind, that learns how to aggregate an ensemble of predictions across languages made with machine translations, which, combined with our SMALR architecture, results in a total improvement over the state-of-the-art by 3-4\%.
\end{itemize}

\section{Related Work}
Transformer~\cite{vaswani_transformer} based representation learning models have become increasingly popular since the release of BERT~\cite{bert}. BERT transfers surprisingly well to other languages, despite having no multilingual training data or explicit multilingual loss~\cite{wu2019beto}. However,~\cite{pires2019multilingual} demonstrates that there is an unequal transfer between different language pairs, notably those with typological differences to English. Both BERT and M-BERT, its multilingual extension, have been shown to be dependent on the number of parameters in the model, which reaches 110M parameters for the smaller base model~\cite{k2019crosslingual}. Thus, as also shown in~\cite{aharoni-etal-2019-massively}, a large number of additional parameters is needed to counter the performance degradation caused by training with many languages. Using the better performing large BERT model is impractical for many vision-language tasks as it contains 340M parameters, leaving little room in many GPUs memory for anything else.

Along with language-only BERT variants, a burst of multimodal BERT-like models have been designed specifically for vision-language tasks~\cite{li2019visualbert,lu2019vilbert,su2019vlbert,tan2019lxmert}. More traditional word embedding models have also been designed for multimodal tasks with the use of either visual-word co-occurrence frequencies~\cite{guptaEmbeddings2019}, multi-task training~\cite{Nguyen_2019_CVPR}, or both~\cite{burnsLanguage2019}, and require significantly less training data to reach similar performance. While these efforts evaluate on many multimodal tasks such as Visual Question Answering~\cite{AntolICCV2015}, Visual Commonsense Reasoning~\cite{VCR}, Phrase Grounding~\cite{plummer2015flickr30k}, and more, they only train and evaluate on a single language.

Recently, several multilingual methods have shown better performance on vision-language tasks than complicated transformer-based methods. LIWE~\cite{Wehrmann_2019_ICCV} is a light-weight character embedding model that can represent many languages with few model parameters. LIWE uses a bidirectional gated recurrent unit (GRU)~\cite{gru} to aggregate 24-D character embeddings for a text query that is encouraged to closely embed semantically similar images and sentences in other languages.   Although LIWE represents a single language well, it suffers from significant performance loss when co-training on multiple languages as shown in Figure~\ref{fig:motivation}(b). Gella~\etal~\cite{gellaEMNLP2017} learns how to relate an image to language-specific representations and also constrain semantically similar sentences across languages to embed nearby each other.  MULE~\cite{kimMULEAAAI2020} learns a universal language embedding so that it can use a single language branch in the multimodal model, significantly reducing the number of parameters required to represent each language compared to Gella~\etal. In addition, MULE combined the same cross-lingual constraints used in both  Gella~\etal and LIWE with an adversarial language classifier to further encourage alignment across languages.  This results in a model that slightly improves performance as more languages are added as shown Figure~\ref{fig:motivation}(b).  However, MULE learns a word-level embedding that requires significantly more parameters than LIWE (approximately 8x more with ten languages), and thus capacity concerns remain when scaling to many languages.
\section{Scalable Multilingual Aligned Language Representation}
\label{sec:SMALR}

In this section we describe how we train our Scalable Multilingual Aligned Language Representation (SMALR) to bridge the gap between scalability and downstream vision-language task performance.  To accomplish this, we assume we are provided with an image and sentences that describe it in multiple languages.  The intuition behind our model is to first learn a universal language embedding which represents all languages, and then learn to relate it to corresponding images using a multimodal model.  In our experiments our multimodal model uses a modified version~\cite{kimMULEAAAI2020} of the Embedding Network architecture~\cite{wang2018learning}, although our approach can be easily adapted to use other multimodal models.  After obtaining image and sentence features, the Embedding Network uses two branches, one for each modality, and projects them into a joint semantic space where distances are meaningful.  The image branch consists of two fully connected layers, while the language branch obtains a sentence representation by passing the final hidden state of a GRU through a fully connected layer.  

Our approach is architecturally similar to MULE~\cite{kimMULEAAAI2020}, but with notable distinctions.  First, MULE learned a unique word embedding for every word in every language (\ie, no shared tokens), whereas we learn an efficient universal embedding with our Hybrid Embedding Model (HEM) that consists of a mix of language-agnostic and language-specific word representations (Section~\ref{subsec:HEM}).  Then, we align our language representations using a novel Masked Cross-Language Model (MCLM) (Section~\ref{subsec:MCLM}) on both the input of the multimodal model and the final language representation of the multimodal model. This acts to supplement the neighborhood constraints, adversarial language classifier, and image-sentence matching losses used by MULE that we briefly review in Section~\ref{subsec:MVSA}.  Finally, we also propose a Cross-Lingual Consistency (CLC) module that boosts model performance in downstream vision-language tasks using machine translation (Section~\ref{subsec:MLI}).  See Fig.~\ref{fig:model} for an overview of our approach.

\subsection{Efficient Multilingual Learning with a Hybrid Embedding Model}
\label{subsec:HEM}

A significant challenge in multilingual representation learning is scaling to many languages, especially when there is a wide disparity in the available training data of different languages.  This is more apparent for vision-language tasks where annotations are very expensive to collect, making it more difficult to learn a good visually-semantically aligned language representation than in monolingual settings~\cite{burnsLanguage2019,li2019visualbert}.  Inspired by work in low-resource neural machine translation~\cite{gu-etal-2018-universal}, we propose a Hybrid Embedding Model (HEM) which projects low-frequency words across languages into a shared latent vocabulary, while allowing the top-$K$ most frequent words in each language to maintain their own unique (language-specific) representation.  The output of the HEM is the universal language embedding that is used as input to the multimodal model in Fig.~\ref{fig:model} and is also used in the language alignment losses (Section~\ref{subsec:MCLM} and Section~\ref{subsec:MVSA}). The value of $K$ can be determined experimentally for any targeted downstream task; we use $K=5000$.

The language-specific word embeddings used for common words roughly follow the implementation used in prior work~\cite{gu-etal-2018-universal,kimMULEAAAI2020}.  We begin by using a monolingual pretrained FastText embedding~\cite{conneau2017word} that has been reduced from 300-D to 50-D using Principal Component Analysis (PCA)~\cite{MACKIEWICZ1993303}. These reduced features are used as input to a fully connected layer that projects them into a 512-D universal embedding space that we align across languages; the alignment is applied with the language-agnostic representations as well (see Section~\ref{subsec:MCLM} and~\ref{subsec:MVSA} for details on our language alignment procedures).

While our language-agnostic representation is similar to Gu~\etal~\cite{gu-etal-2018-universal}, it has some key differences.  Specifically, Gu~\etal project all words into a universal embedding space with learned language-specific mappings. A soft-attention module is used over the universal embedding features (as it assumes an aligned cross-lingual input)  to obtain mixing weights; these weights are then used to combine the language-agnostic features. While this does enable feature sharing across languages, it does not reduce the number of trainable parameters in the network, as a language-specific representation is still necessary for all words in the vocabulary. Additionally, aggregating the features in the latent vocabulary using soft-attention weights per-word is costly, especially for large vocabularies. Instead, we perform a pretraining step where we learn both the initial representation of the latent vocabulary as well as how to assign the infrequent words to entries in it. We use a hard attention mechanism that is directly predicted from FastText features, in which each vocabulary word is mapped to only a single language-agnostic token, as opposed to an interpolation of many. This allows us to avoid both computing a language-specific representation for the uncommon words and aggregating the latent vocabulary features on a per-word basis.

To obtain our latent shared vocabulary in the pretraining step, we learn to embed semantically similar sentences near each other using a triplet loss.  More formally, given a triplet of items $(x, y^+, y^-)$ that can be decomposed into a positive pair $(x, y^+)$ and a negative pair $(x, y^-)$, a triplet loss is computed as:
\begin{equation}
    L_{triplet}(x, y^+, y^-) = \max(0, m + d(x, y^+) - d(x, y^-))
    \label{eq:triplet}
\end{equation}
\noindent where $d(x, y)$ is a distance function, and $m$ is a scalar parameter.  We use cosine distance for all triplet losses and set $m=0.05$.  Following the methodology of~\cite{kimMULEAAAI2020,wang2018learning}, we construct minibatches by providing semantically similar sentence pairs as input and consider any non-paired sentence as a negative example. These negatives are randomly sampled from each minibatch. We enumerate all triplets in the minibatch and compute the loss over the top-$N$ most violated constraints, where $N=10$ in our experiments. Note that these sentences may not come from the same language, so sentences referring to the same image in different languages are also used as positive pairs. To predict which latent embedding we map a source word to, we use sentence representations obtained by feeding FastText embeddings into a fully connected layer. With this mapping, we average the latent embeddings of each word for use in Eq. (1) during the pretraining step, which has been shown to be an efficient, high-performing representation~\cite{arora,burnsLanguage2019}.

Instead of deterministically mapping to the latent token which achieves the best score, we randomly choose from the top $M$ scoring tokens with probability $p$, which we refer to as exploration parameters. This helps ensure that spurious mappings are not learned, typically resulting in a 2\% performance improvement (see supplementary for a detailed comparison).
While we freeze the latent token assignments when training the full model, we allow the features themselves to be fine-tuned. Our experiments use a latent vocabulary size of $40K$ tokens, with exploration parameters $p = 0.2$, $M = 20$.  In practice not all latent tokens are used at the end of pretraining; these are dropped when training the full model. 

\subsection{Masked Cross-Language Modeling (MCLM)}
\label{subsec:MCLM}
Masked Language Modeling has proven to be useful in training language representations by masking some tokens of an input sentence and then trying to predict the missing tokens~\cite{bert}. We present a generalization of this approach to a multilingual scenario to encourage stronger cross-language alignment. In MCLM, we assume we have paired sentences across different languages.  These sentences need not be direct translations of each other, but, as our experiments will show, they simply need to be semantically related.  This is important as vision-language datasets do not always have paired text queries that are direct translations, but are often independently generated instead (\eg~\cite{elliott2016multi30k,miyazaki2016cross,li2019coco}).

Traditional Masked Language Modeling makes predictions about a single masked token using its surrounding words as context. The words immediately surrounding a token referring to the same entity between sentences in different languages may vary greatly due to differences in grammar. Thus, even using a dictionary between languages to identify word correspondences may not provide useful context. Instead, we use the intuition that semantically similar sentences should contain comparable information across languages, so a sentence in one language could be used as context to predict missing information in another. Conneau~\etal~\cite{conneaumclm} similarly use masking for improved language alignment. However, our approach does not require parallel data and may sample amongst any of the languages. 
Lastly, unlike~\cite{conneaumclm} which computes its loss on the predicted word, our objective in Eq. (2) is computed on the fully reconstructed sentences.

More formally, for a pair of languages $(i,j)$, we obtain sentences $(S_i, S_j)$ such that both sentences describe the same image (\ie, they are semantically similar to each other). Then, we randomly replace some portion of their words with a special MASK token to obtain masked representations $(S^{m}_{i}, S^{m}_{j})$. These are concatenated together and fed into a fully connected layer that is shared across language pairs to predict the missing information in both sentences $(S^{'}_{i}, S^{'}_{j})$. Our MCLM loss then compares this to the unmasked sentences, \ie, 
\begin{equation}
    L_{mask} = ||\ell_2(S^{m}_{i} + S^{'}_{i}) - \ell_2(S_i)|| + ||\ell_2(S^{m}_{j} + S^{'}_{j}) - \ell_2(S_j)||,
    \label{eq:masking_loss}
\end{equation}
\noindent where $\ell_2$ identifies vectors forced to have unit norm. Both average embedding and LSTM representations are used; details can be found in the supplementary. We compute the masking loss in Eq.~(\ref{eq:masking_loss}) for all unique pairs of languages in our experiments, and found masking 20\% of the words in the sentences worked best. 

\subsection{Multilingual Visual-Semantic Alignment}
\label{subsec:MVSA}

In this section we briefly review the visual-semantic alignment constraints used by MULE~\cite{kimMULEAAAI2020} that we also employ.  First, we use neighborhood constraints~\cite{wang2018learning} that we shall refer to as $L_{nc}$ to encourage similar sentences to embed nearby each other using a triplet loss (\ie, Eq.~(\ref{eq:triplet})).  Just as with the MCLM module described in Section~\ref{subsec:MCLM}, these neighborhood constraints are applied to both the universal language embedding (\ie, the output of the HEM module) as well as the final language representation from the multimodal model as shown in Fig.~\ref{fig:model}.  The second component of the MULE alignment constraint consists of an adversarial language classifier. We shall refer to this classifier loss as $L_{adv}$, using the approach of~\cite{kimMULEAAAI2020}, whose goal is to ensure that the representations of the different languages in the universal embedding have similar feature distributions.  The last component of the MULE constraint is used to train the multimodal model to embed the images and sentences near each other using a triplet loss.  This uses a bidirectional triplet loss function, \ie, for image $I$ and paired sentences $(Q^+, Q^-)$ representing a positive and negative sentence pair, respectively, and sentence $Q$ and its paired images $(I^+, I^-)$, this multimodal loss would be,

\begin{equation}
    L_{mm} = L_{triplet}(I, Q^+, Q^-) + \lambda_1 L_{triplet}(Q, I^+, I^-)
    \label{eq:mm}
\end{equation}

\noindent where $\lambda_1$ is a scalar parameter, which we set to 1.5 in our experiments. In addition to using the unmasked sentence representations for the multimodal loss, we observe that most sentences retain their overall semantic meaning if you remove just a few words at random.  Using this intuition, we also compute Eq.~(\ref{eq:mm}) using the masked sentences $(S^{m}_{i}, S^{m}_{j})$ from the MCLM module, which we found provides a small, but consistent improvement to performance. As a reminder, all triplet losses use the implementation details (\eg hyperparameter settings and hard-negative mining) as described in the first part of Section~\ref{sec:SMALR}. Our total loss function to train SMALR is then,

\begin{equation}
    L_{SMALR} = L_{mm} + \lambda_2 L_{mask} + \lambda_3 L_{adv} + \lambda_4 L_{nc}
\end{equation}

\noindent where $\lambda_{2-4}$ are scalar parameters that we set to (1e-4, 1e-6, 5e-2), respectively.

\subsection{Cross-Lingual Consistency}
\label{subsec:MLI}

Prior work on multilingual vision-language tasks has primarily focused on how to change training procedures or architectures to support multiple languages, and does not fully take advantage of this multilingual support at test time.  In particular, we argue that semantically similar sentences in different languages may capture complementary information, and therefore, considering the predictions made in other languages may improve performance.  We validate our intuition by obtaining machine translations of a query in the other languages supported by our model.  More formally, suppose we have a set of languages $L$. Given a query $q$ in language $l_i \in L$, we translate $q$ to all other supported languages in $L\setminus\{l_i\}$ and use this as input to our Cross-Lingual Consistency (CLC) module.

We propose two variants of CLC: CLC-A and CLC-C.
CLC-A simply averages matching scores over all languages, and does not require any additional parameters. CLC-C, on the other hand, uses a small Multilayer Perceptron (MLP) to aggregate the scores of each language, which enables us to consider the relative information present in each language's predictions. This MLP has two layers with input size $|L|$ and 32 hidden layer units (\ie, it has 352 learnable parameters) and all parameters are initialized with uniform weight. We train the CLC-C module separately to SMALR using the validation set for 30 iterations.  No minibatches are employed (\ie, it is trained with all image-sentence pairs at once) and it is trained using the multimodal triplet loss described in Eq.~(\ref{eq:mm}).

\section{Experimental Setup}
\label{sec:exps}
\noindent\textbf{Datasets.}
SMALR is evaluated on bidirectional retrieval with Multi30K~\cite{barrault2018findings,elliott2017findings,elliott2016multi30k} and MSCOCO~\cite{li2019coco,lin2014microsoft,miyazaki2016cross}. The Multi30K dataset is built off of Flickr30K~\cite{young2014image}, which originally contained 31,783 images and five English descriptions per image.~\cite{barrault2018findings,elliott2017findings,elliott2016multi30k} obtained annotations in German, French, and Czech, resulting in a four-language dataset. Multi30K contains five descriptions per image in English and German, but only one per image in French and Czech; the latter two were collected as human-generated translations of the English annotations. We use the 29K/1K/1K train/test/val splits from the original dataset~\cite{young2014image}. 

MSCOCO is approximately four times the size of Multi30K, with 123,287 images. There are five human-generated captions per image in English, but significantly fewer in Chinese and Japanese. YJ Captions~\cite{miyazaki2016cross} introduced Japanese annotations for MSCOCO, but only provides five captions per image for a subset of about 26K images.~\cite{li2019coco} extended MSCOCO with a total of 22,218 Chinese captions for 20,341 images. We use train/test/validation splits as defined in~\cite{kimMULEAAAI2020}. 

We augment both datasets with machine translations so every image contains at least five sentences for ten languages: English, German, Czech, French, Chinese, Japanese, Arabic, Afrikaans, Korean, and Russian. All models we compare to are trained using this augmented training set.  For languages with no human-generated sentences, we use machine translations at test time as well.  We found using translations at test time did not affect the relative performance of different methods in our experiments. See the supplementary for details.
\smallskip

\smallskip

\noindent\textbf{Visual Features.} 
We use ResNet-152~\cite{He2015} features trained on ImageNet~\cite{imagenet_cvpr09} as input to the Embedding Network (EmbN)~\cite{wang2018learning}. As done in~\cite{kimMULEAAAI2020}, we average visual features over ten 448x448 image crops. This generates an image embedding of size 2048, which is then passed through a pair of fully connected layers. The resulting 512-D embedding can be used in the shared image-sentence embedding space. The learning rate was set to $1 e^{-3}$ for the HEM and LA models; remaining hyperparameters are consistent with those in~\cite{kimMULEAAAI2020}.

Note that all LIWE~\cite{Wehrmann_2019_ICCV} experiments use bottom-up Faster R-CNN~\cite{fasterrcnn} visual features trained on Visual Genome~\cite{krishnavisualgenome}.  This represents a significant increase in the annotation cost compared to our approach, which doesn't use these annotations. Visual Genome also contains MSCOCO~\cite{lin2014microsoft} images, which means that there is train/test contamination, as LIWE's features are extracted using  the pretrained, publicly available model from~\cite{Anderson_2018_CVPR}.
\smallskip

\noindent\textbf{Metrics.}
For our results, we report the mean Recall (mR) across Recall@$K$, with $K\in[1,5,10]$, for both the image-sentence and sentence-image directions per language. All recall values can be found in the supplementary. We also provide an average mR across all languages to serve as a global performance metric: ``A" in Tables~\ref{tab:coco} and~\ref{tab:multi30k}. The human average, ``HA," refers to the average mR over the languages which have human-generated annotations (\ie English, Chinese, and Japanese for MSCOCO, and English, German, French, and Czech for Multi30K).
\smallskip

\noindent\textbf{Comparative Evaluation.} We compare the following methods:
\begin{itemize}
    \item \textbf{Frequency Thresholding:} We drop words that occur fewer than $t$ times in the training set. Results are reported in  Figure~\ref{fig:vocabperf}.
    \item \textbf{PCA Reduction:} We use PCA~\cite{MACKIEWICZ1993303} to reduce the size of the initial 300-D FastText word embeddings. Results are reported in Figure~\ref{fig:vocabperf}.
    \item \textbf{Dictionary Mapping:} We map words that occur fewer than $t$ times in non-English languages to English using dictionaries~\cite{conneau2017word}. By mapping rare words in other languages to English, some information may be lost, but the token will still exist indirectly in the vocabulary. Results are reported in Figure~\ref{fig:vocabperf}.
    \item \textbf{Language-Agnostic (LA):} We compare to only using a latent vocabulary as described in Section~\ref{subsec:HEM} with 40K tokens, \ie not using any language specific features. Results are in Tables~\ref{tab:coco} and~\ref{tab:multi30k}.
    \item \textbf{HEM:} We evaluate our full hybrid embedding model (Section~\ref{subsec:HEM}), which uses a mix of language-agnostic and language-specific representations. This baseline does not include MCLM nor CLC. Results are in Tables~\ref{tab:coco} and~\ref{tab:multi30k}.
    \item \textbf{SMALR:} Our base SMALR is composed of the HEM (Section~\ref{subsec:HEM}) and MCLM (Section~\ref{subsec:MCLM}) components of our model. We compare to our complete SMALR which makes use of CLC variants (CLC-A and CLC-C, described in Section~\ref{subsec:MLI}). Results are in Tables~\ref{tab:coco} and~\ref{tab:multi30k}.
\end{itemize}

\noindent Note that the first line of Tables~\ref{tab:coco} and~\ref{tab:multi30k}, \textbf{Trans To En}, refers to using machine translation on non-English sentences, and then using an English-only trained Embedding Network~\cite{wang2018learning}, providing a strong baseline method to compare to.

\section{Multilingual Image-Sentence Retrieval Results}
\begin{figure}[t]
    \includegraphics[width=6cm]{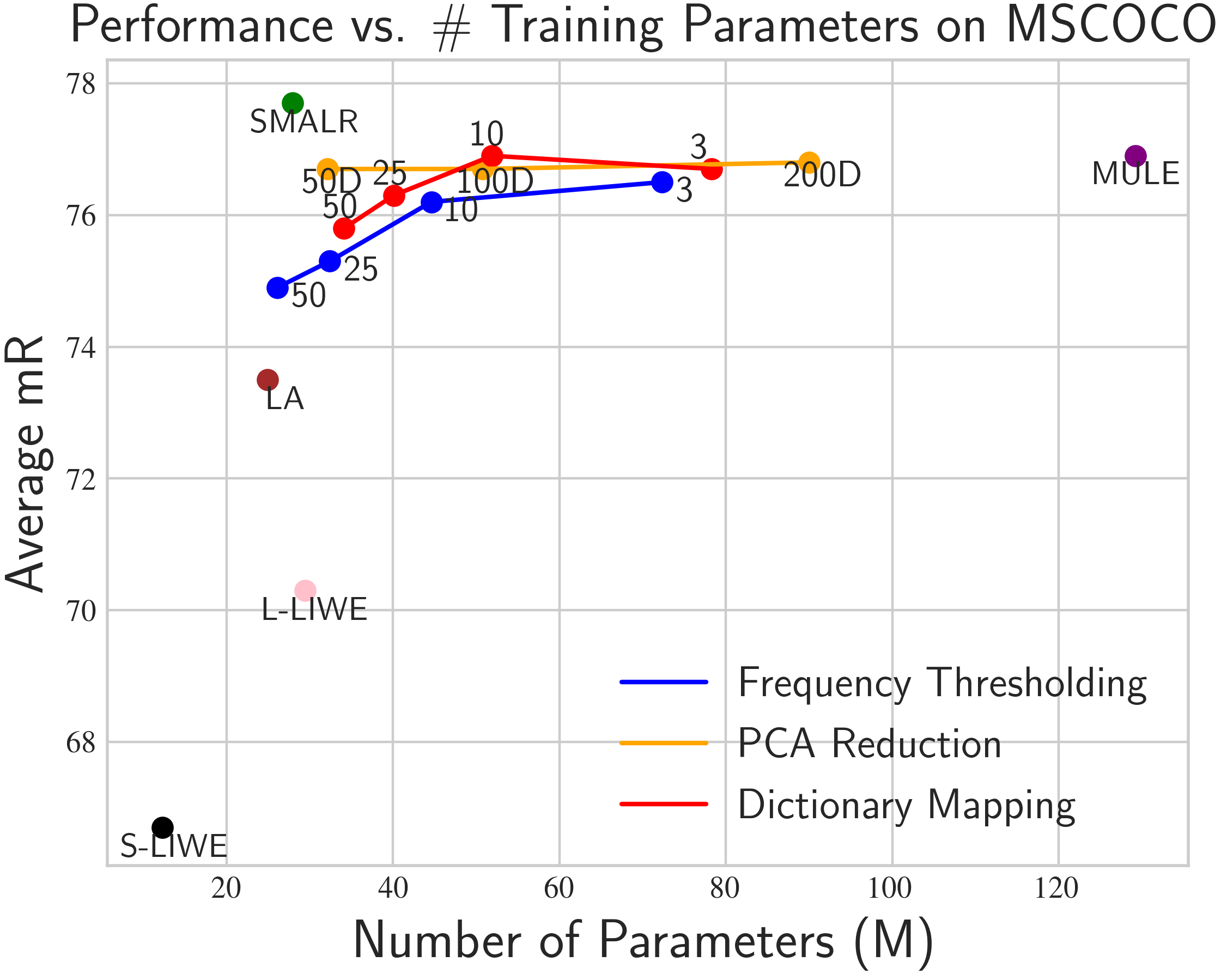}
    \includegraphics[width=5.9cm]{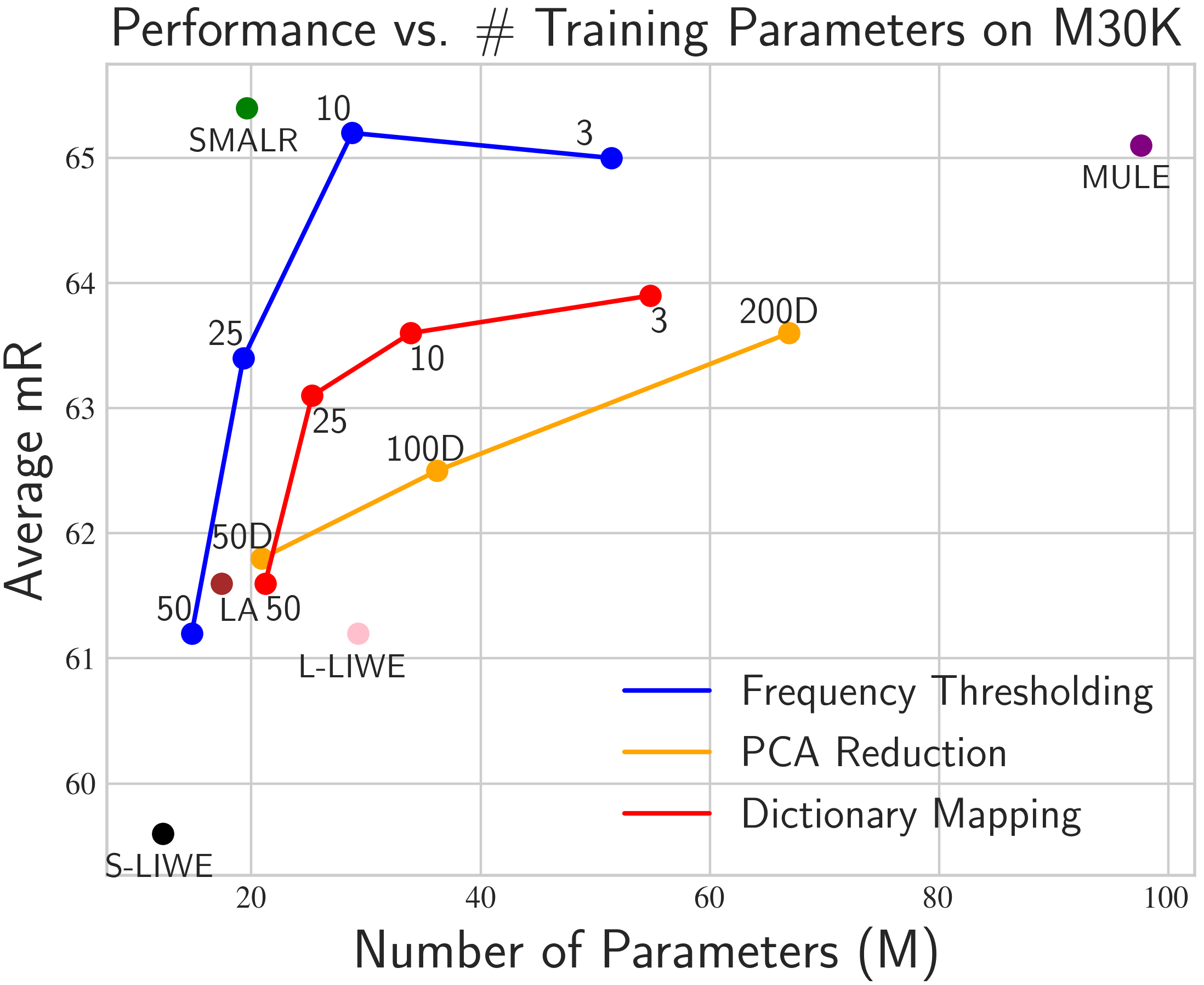}
    \caption{We compare three types of vocabulary reduction: frequency thresholding, PCA dimensionality reduction, and mapping rare words to English with the use of dictionaries. The left-hand side evaluates on MSCOCO, the right on Multi30K. We have additional standalone points for the small LIWE (S-LIWE), large LIWE (L-LIWE), MULE, language agnostic vocabulary (LA), and our model, SMALR}
    \label{fig:vocabperf}
\end{figure}

We provide results for MSCOCO and Multi30K in Table~\ref{tab:coco} and Table~\ref{tab:multi30k}, respectively, which contain comparisons to prior work on fewer languages \textbf{(a)}, adaptations of prior work to our setting \textbf{(b)}, and our model variants \textbf{(c)}. SMALR obtains consistent performance gains when evaluating on ten languages over the state-of-the-art (S-LIWE, line 3\textbf{(b)}) while also being more efficient than high-performing models like MULE (line 5\textbf{(b)}). SMALR outperforms S-LIWE by 11 points on MSCOCO and 5.8 points on Multi30K (line 3\textbf{(c)} versus 3\textbf{(b)}). A parameter comparison is later shown in Figure~\ref{fig:vocabperf}. SMALR's initial Language-Agnostic (LA) baseline alone is able to boost performance over previous scalable method LIWE by 2-7 points. The HEM, which combines language-agnostic and language-specific embeddings as described in Section~\ref{subsec:HEM}, consistently improves upon the fully language-agnostic vocabulary, even though they share the same latent vocabulary size of 40K tokens. This points to the utility of our hybrid embedding space, which improves performance upon LA by 3.4 average mR on MSCOCO and 2.4 average mR on Multi30K while adding only a few parameters. 

When MCLM losses are added, referred to as SMALR in Tables~\ref{tab:coco} and~\ref{tab:multi30k} (line 3\textbf{(c)}), mR improves for nearly all languages. This is significant, because we find more compact models like LIWE degrade with additional languages when using the same number of parameters (S-LIWE). The LA baseline is still able to outperform L-LIWE on MSCOCO and Multi30K, in which LIWE learns an embedding five fold larger to try to compensate for the increased number and diversity of languages (120-D instead of 24-D embedding). This suggests that the masking process may help regain some semantic information that is lost when tokens are mapped to the language-agnostic space.

We next evaluate two CLC variants that use machine translations at test time (described in Section~\ref{subsec:MLI}) on top of SMALR: an average ensemble over all languages (CLC-A), and a weighted ensemble which makes use of a simple classifier (CLC-C).
CLC-A uses no additional test-time parameters, and increases the human average performance by 1-3 points, with a larger gain on Multi30K. This may be because more languages can be leveraged on Multi30K (four versus three, compared to MSCOCO). Surprisingly, English performance improves the most amongst CLC-A metrics on Multi30K, demonstrating that certain image-sentence pairs can be better retrieved from the queries in other languages, which may better capture the visual semantics of the same image. CLC-C further improves the human average over CLC-A by 0.9 points on MSCOCO and 0.5 points on Multi30K, using negligible additional parameters.
\smallskip

\begin{table}[t]
\setlength{\tabcolsep}{1.pt}
\begin{center}
\caption{MSCOCO multilingual bidirectional retrieval results.  (a) contains results from prior work, (b) contains reproductions of two state-of-the art methods evaluated for our scenario using their code, and (c) contains variants of our model}
\label{tab:coco}
\begin{tabular}{|rl|c|c|c|c|c|c|c|c|c|c|c|c|}
\hline
& Model & En & De\footnotemark[1] & Fr\footnotemark[1] & Cs\footnotemark[1] & Cn & Ja & Ar\footnotemark[1] & Af\footnotemark[1] & Ko\footnotemark[1] & Ru\footnotemark[1] & HA & A\\
\hline
\hline
\textbf{(a)} & Trans. to En~\cite{kimMULEAAAI2020} & 75.6 & -- & -- & -- & 72.2 & 66.1 & -- &-- & -- &-- & 71.3 & --\\
& EmbN~\cite{wang2018learning} & 76.8 & -- & -- & -- & 73.5 & 73.2 & -- &-- &-- & -- & 74.5 & -- \\
& PAR. EmbN~\cite{gellaEMNLP2017} & 78.3 & -- & -- & -- & 73.5 & 76.0 & -- & -- & -- & -- & 75.9 & -- \\
& MULE~\cite{kimMULEAAAI2020} & 79.5 & -- & -- & -- & 74.8 & 76.3 & -- & -- & -- & -- & 76.9 & --\\
\hline
\hline
 
 \textbf{(b)} & (1) S-LIWE~\cite{Wehrmann_2019_ICCV}\footnotemark[2] & 80.9 & -- & -- & -- & -- & 73.6 & -- & -- & -- & -- & -- & -- \\
 & (2) S-LIWE\footnotemark[2] & 77.4 &-- & -- & --& --& 66.6& -- & -- & -- & -- & -- & -- \\
  & (10) S-LIWE\footnotemark[2] & 77.3 & 67.4 & 68.5 & 66.9 & 64.5 & 65.8 & 63.8 & 66.2 & 63.1 & 63.6 & 69.2 & 66.7\\
  & (10) L-LIWE\footnotemark[2] & 79.1 & 71.2 & 70.3 & 70.1 & 70.0& 69.6 & 67.5 & 68.9 & 66.2 & 69.6& 72.9 & 70.3\\
& MULE~\cite{kimMULEAAAI2020} & 79.0 & 77.2 & 76.8 & 77.8 & 75.6 & 75.9 & 77.2 & 77.8 & 74.3 & 77.3 & 76.8 & 76.9\\
\hline\hline
\textbf{(c)} & Language-Agnostic & 75.0 & 74.3 & 74.1 & 73.4 & 72.3 & 72.1 & 74.4 & 74.7 & 71.6 & 72.7 & 73.1 & 73.5\\
& HEM & 78.7 & 77.3 & 76.4 & 77.9 & 76.7 & 76.3 & 77.0 & 76.7 & \textbf{75.5} & 77.0 & 77.3 & 76.9\\
& SMALR  & 79.3 & \textbf{78.4} & \textbf{77.8} &\textbf{78.6} & 76.7 & 77.2 & \textbf{77.9}& \textbf{78.2} & 75.1 & \textbf{78.0} & 77.7 & \textbf{77.7}\\
& SMALR-CLC-A & 81.2 & -- & -- & -- & 79.6 & 75.0 & -- & -- & -- & -- & 78.6 &--\\
& SMALR-CLC-C & \textbf{81.5} & -- & -- & -- & \textbf{80.1} & \textbf{77.5} & -- & -- & -- & -- & \textbf{79.7}&--\\
\hline
\end{tabular}
\end{center}
\footnotemark[1]uses translations from English for testing\\ 
\footnotemark[2]visual features trained using outside dataset that includes some test images
\end{table}

\begin{table}[t]
\setlength{\tabcolsep}{1.pt}
\begin{center}
\caption{Multi30K multilingual bidirectional retrieval results. (a) contains results from prior work, (b) contains reproductions of two state-of-the art methods evaluated for our scenario using their code, and (c) contains variants of our model}
\label{tab:multi30k}
\begin{tabular}{|rl|c|c|c|c|c|c|c|c|c|c|c|c|}
\hline
& Model & En & De & Fr & Cs & Cn\footnotemark[1] & Ja\footnotemark[1] & Ar\footnotemark[1] & Af\footnotemark[1] & Ko\footnotemark[1] & Ru\footnotemark[1] & HA & A\\
\hline
\hline
\textbf{(a)} & Trans. to En~\cite{kimMULEAAAI2020} & 71.1 & 48.5 & 46.7 & 46.9 & -- & -- & -- &-- & -- &-- & 53.3 & --\\
& EmbN~\cite{wang2018learning} & 72.0 & 60.3 & 54.8 & 46.3 & -- & --& -- &-- &-- & -- & 58.4 & -- \\
& PAR. EmbN~\cite{gellaEMNLP2017} & 69.0 & 62.6 & 60.6 & 54.1 & -- & -- & -- & -- & -- & -- & 61.6 & -- \\
& MULE~\cite{kimMULEAAAI2020} & 70.3 & 64.1 & 62.3 & 57.7 & -- & -- & -- & -- & -- & -- & 63.6 & --\\
\hline
\hline

\textbf{(b)} & (1) S-LIWE~\cite{Wehrmann_2019_ICCV}\footnotemark[2] & \textbf{76.3} & \textbf{72.1}& -- & -- & -- & -- & -- & -- & -- & -- & -- & --\\
& (2) S-LIWE\footnotemark[2] & 75.6 & 66.1 & -- & -- & -- & -- & -- & -- & -- & -- & -- & --\\
& (10) S-LIWE\footnotemark[2] & 75.2 & 65.2 & 51.8 & 50.0 & 54.1 & 56.2 & 62.7 & 62.8 & 54.5 & 63.1& 60.6 & 59.6\\
& (10) L-LIWE\footnotemark[2] & 75.9 & 66.7 & 53.3 & 51.3 & 56.9 & 56.3 & 65.0 & 63.7& 57.1 & 65.4 & 61.9 & 61.2 \\
& MULE~\cite{kimMULEAAAI2020} & 70.7 & 63.6 & 63.4 & 59.4 & \textbf{64.2} & \textbf{67.3} & 65.8 & 67.3 & 63.6 & 65.4 & 64.3 & 65.1\\
\hline\hline
\textbf{(c)} & Language-Agnostic & 65.5 & 61.3 & 59.9 & 54.0 & 59.4 & 64.7 & 63.9 & 66.5 & 60.3 & 60.3 & 60.2 & 61.6 \\
& HEM & 69.2 & 62.8 & 63.3 & 60.0 & 62.4 & 66.3 & 64.5 & 66.8 & 62.3 & 62.6 & 63.8 & 64.0\\
& SMALR & 69.6 & 64.7 & 64.5 & 61.1 & 64.0 & 66.7 & \textbf{66.0} & \textbf{67.4} & \textbf{64.2} & \textbf{65.7} & 65.0 & \textbf{65.4}\\
& SMALR-CLC-A & 74.1 & 68.9 & 65.2 & 64.5 & -- & -- & -- & -- & -- & -- & 68.2 &--\\
& SMALR-CLC-C & 74.5  & 69.8& \textbf{65.9} & \textbf{64.8} & -- & -- & -- & -- & -- & -- & \textbf{68.7} &--\\
\hline
\end{tabular}
\end{center}
\footnotemark[1]uses translations from English for testing\\ \footnotemark[2]visual features trained using outside dataset
\end{table}

\noindent\textbf{Parameter reduction method comparison}. We present a comparison of baseline vocabulary reduction techniques, described in Section~\ref{sec:exps}, against prior works LIWE and MULE, in addition to our method SMALR (consisting of only HEM and MCLM components in Figure~\ref{fig:vocabperf}). The frequency thresholding and dictionary mapping labels represent the threshold with which we drop infrequent words or map them to English (\eg the blue 50 data point represents dropping words that occur fewer than 50 times). PCA point labels represent the dimensionality we reduce our input vectors to (\eg 300D $\rightarrow$ 50D, 100D, or 200D). 

In our comparison of vocabulary reduction methods, frequency thresholding with $t=50$ and vanilla language-agnostic vocabularies (LA) obtain better performance than both LIWE variants on Multi30K, without adding significantly more parameters, as shown on the right of Figure~\ref{fig:vocabperf}. While more model parameters are needed for MSCOCO, due to the increased vocabulary size, all baselines and prior work MULE significantly outperform LIWE. This demonstrates that more-complex character-based models do not necessarily obtain competitive performance with few parameters when addressing a larger multilingual scenario.

SMALR outperforms all baselines for MSCOCO, as seen on the left of Figure~\ref{fig:vocabperf}, outperforming S-LIWE by over 10 points and using fewer parameters than L-LIWE. We also find that average mean recall performance on MSCOCO is more robust to vocabulary reduction, with a maximum range of about 1.5 average mR between the most extreme reduction and the least. We believe this may be due to the size discrepancy between the two datasets, as MSCOCO is approximately four times the size of Multi30K. PCA reduction appears to have a more linear effect as parameters increase on both datasets. Since Multi30K performance is more sensitive to the number of parameters, it is significant that our SMALR model, in green, (which does not yet make use of our cross-lingual consistency module in Figure~\ref{fig:vocabperf}) outperforms all other models while having less than 20M parameters, 1/5th the parameter count of high performing MULE.

In addition to SMALR outperforming MULE on both datasets while using significantly fewer trainable parameters, we find MULE even fails to outperform simple baselines such as dictionary mapping on MSCOCO. This exposes that the large number of parameters used in MULE are unnecessary for performance gains. While SMALR uses more parameters during training than S-LIWE, we have far fewer test-time parameters. We reduce the computation needed for evaluation by using precomputed language representations from training. This reduces the entire SMALR model to the image-sentence matching model with our CLC add-on, totaling only 7.1M parameters, now fewer than S-LIWE.

\section{Conclusion}
We have presented a Scalable Multilingual Aligned Representation (SMALR), which addresses the trade-off between multilingual model size and downstream vision-language task performance. Our approach is modular, and thus can be used as a drop-in language representation for any vision-language method/task. SMALR outperforms all prior work on the task of multilingual image-sentence retrieval on average across ten diverse languages, with the use of a hybrid embedding model, masked cross-language modeling loss, and cross-lingual consistency module. Our hybrid embedding model significantly reduces the input to a language model by mapping most tokens to a fixed size, shared vocabulary. The masking procedure aligns our diverse set of languages and uses the multimodal model to provide additional alignment with visual grounding. We find that both cross-lingual consistency modules better aggregates retrieved results, boosting performance with minimal additional parameters. This is all accomplished with less than 20M trainable parameters, significantly reducing oversized prior work by 1/5th, while improving performance over the state-of-the-art by 3-4\%. 
\smallskip

\noindent\textbf{Acknowledgements}
This work is funded in part by the NSF, DARPA LwLL, and DARPA XAI grants, including NSF grant 1838193.
%
%
\bibliographystyle{splncs04}
\bibliography{egbib}

\begin{thebibliography}{10}
\providecommand{\url}[1]{\texttt{#1}}
\providecommand{\urlprefix}{URL }
\providecommand{\doi}[1]{https://doi.org/#1}

\bibitem{aharoni-etal-2019-massively}
Aharoni, R., Johnson, M., Firat, O.: Massively multilingual neural machine
  translation. In: Proceedings of the 2019 Conference of the North {A}merican
  Chapter of the Association for Computational Linguistics: Human Language
  Technologies, Volume 1 (Long and Short Papers) (Jun 2019)

\bibitem{Anderson_2018_CVPR}
Anderson, P., He, X., Buehler, C., Teney, D., Johnson, M., Gould, S., Zhang,
  L.: Bottom-up and top-down attention for image captioning and visual question
  answering. In: The IEEE Conference on Computer Vision and Pattern Recognition
  (CVPR) (2018)

\bibitem{AntolICCV2015}
Antol, S., Agrawal, A., Lu, J., Mitchell, M., Batra, D., Zitnick, C.L., Parikh,
  D.: {VQA: Visual Question Answering}. In: The IEEE International Conference
  on Computer Vision (ICCV) (2015)

\bibitem{arora}
Arora, S., Liang, Y., Ma, T.: A simple but tough-to-beat baseline for sentence
  embeddings. In: International Conference on Learning Representations (ICLR)
  (2017)

\bibitem{bilingual2}
Artetxe, M., Labaka, G., Agirre, E.: Learning principled bilingual mappings of
  word embeddings while preserving monolingual invariance. In: Empirical
  Methods in Natural Language Processing (EMNLP). pp. 2289--2294 (2016)

\bibitem{barrault2018findings}
Barrault, L., Bougares, F., Specia, L., Lala, C., Elliott, D., Frank, S.:
  Findings of the third shared task on multimodal machine translation. In:
  Proceedings of the Third Conference on Machine Translation: Shared Task
  Papers. pp. 304--323 (2018)

\bibitem{bojanowski2017enriching}
Bojanowski, P., Grave, E., Joulin, A., Mikolov, T.: Enriching word vectors with
  subword information. Transactions of the Association for Computational
  Linguistics (TACL)  \textbf{5},  135--146 (2017)

\bibitem{burnsLanguage2019}
Burns, A., Tan, R., Saenko, K., Sclaroff, S., Plummer, B.A.: Language features
  matter: Effective language representations for vision-language tasks. In: The
  IEEE International Conference on Computer Vision (ICCV) (2019)

\bibitem{gru}
Cho, K., van Merrienboer, B., G{\"{u}}l{\c{c}}ehre, {\c{C}}., Bougares, F.,
  Schwenk, H., Bengio, Y.: Learning phrase representations using {RNN}
  encoder-decoder for statistical machine translation. In: Empirical Methods in
  Natural Language Processing (EMNLP) (2014)

\bibitem{conneaumclm}
Conneau, A., Lample, G.: Cross-lingual language model pretraining. In: Advances
  in Neural Information Processing Systems (NeurIPS) (2019)

\bibitem{conneau2017word}
Conneau, A., Lample, G., Ranzato, M., Denoyer, L., J{\'e}gou, H.: Word
  translation without parallel data. In: International Conference on Learning
  Representations (ICLR) (2018)

\bibitem{imagenet_cvpr09}
Deng, J., Dong, W., Socher, R., Li, L.J., Li, K., Fei-Fei, L.: {ImageNet: A
  Large-Scale Hierarchical Image Database}. In: The IEEE Conference on Computer
  Vision and Pattern Recognition (CVPR) (2009)

\bibitem{bert}
Devlin, J., Chang, M.W., Lee, K., Toutanova, K.: Bert: Pre-training of deep
  bidirectional transformers for language understanding. In: arXiv:1810.04805v1
  (2018)

\bibitem{elliott2017findings}
Elliott, D., Frank, S., Barrault, L., Bougares, F., Specia, L.: Findings of the
  second shared task on multimodal machine translation and multilingual image
  description. arXiv:1710.07177  (2017)

\bibitem{elliott2016multi30k}
Elliott, D., Frank, S., Sima'an, K., Specia, L.: Multi30k: Multilingual
  english-german image descriptions. arXiv:1605.00459  (2016)

\bibitem{gellaEMNLP2017}
Gella, S., Sennrich, R., Keller, F., Lapata, M.: Image pivoting for learning
  multilingual multimodal representations. In: Empirical Methods in Natural
  Language Processing (EMNLP) (2017)

\bibitem{balanced_vqa_v2}
Goyal, Y., Khot, T., Summers{-}Stay, D., Batra, D., Parikh, D.: Making the {V}
  in {VQA} matter: Elevating the role of image understanding in {V}isual
  {Q}uestion {A}nswering. In: The IEEE Conference on Computer Vision and
  Pattern Recognition (CVPR) (2017)

\bibitem{gu-etal-2018-universal}
Gu, J., Hassan, H., Devlin, J., Li, V.O.: Universal neural machine translation
  for extremely low resource languages. In: Proceedings of the 2018 Conference
  of the North {A}merican Chapter of the Association for Computational
  Linguistics: Human Language Technologies (ACL-HLT) (2018)

\bibitem{guptaEmbeddings2019}
Gupta, T., Schwing, A., Hoiem, D.: Vico: Word embeddings from visual
  co-occurrences. In: The IEEE International Conference on Computer Vision
  (ICCV) (2019)

\bibitem{He2015}
He, K., Zhang, X., Ren, S., Sun, J.: Deep residual learning for image
  recognition. arXiv:1512.03385  (2015)

\bibitem{k2019crosslingual}
K, K., Wang, Z., Mayhew, S., Roth, D.: Cross-lingual ability of multilingual
  bert: An empirical study. arXiv:1912.07840  (2019)

\bibitem{kazemzadeh-EtAl:2014:EMNLP2014}
Kazemzadeh, S., Ordonez, V., Matten, M., Berg, T.: Referitgame: Referring to
  objects in photographs of natural scenes. In: Empirical Methods in Natural
  Language Processing (EMNLP) (2014)

\bibitem{kimMULEAAAI2020}
Kim, D., Saito, K., Saenko, K., Sclaroff, S., Plummer, B.A.: Mule: Multimodal
  universal language embedding. In: AAAI Conference on Artificial Intelligence
  (2020)

\bibitem{klein2014fisher}
Klein, B., Lev, G., Sadeh, G., Wolf, L.: Fisher vectors derived from hybrid
  gaussian-laplacian mixture models for image annotation. In: The IEEE
  Conference on Computer Vision and Pattern Recognition (CVPR) (2015)

\bibitem{krishnavisualgenome}
Krishna, R., Zhu, Y., Groth, O., Johnson, J., Hata, K., Kravitz, J., Chen, S.,
  Kalantidis, Y., Li, L.J., Shamma, D.A., Bernstein, M., Fei-Fei, L.: Visual
  genome: Connecting language and vision using crowdsourced dense image
  annotations. International Journal of Computer Vision (IJCV)  (2017)

\bibitem{li2019visualbert}
Li, L.H., Yatskar, M., Yin, D., Hsieh, C.J., Chang, K.W.: Visualbert: A simple
  and performant baseline for vision and language. arXiv:1908.03557  (2019)

\bibitem{li2019coco}
Li, X., Xu, C., Wang, X., Lan, W., Jia, Z., Yang, G., Xu, J.: Coco-cn for
  cross-lingual image tagging, captioning and retrieval. IEEE Transactions on
  Multimedia  (2019)

\bibitem{lin2014microsoft}
Lin, T.Y., Maire, M., Belongie, S., Hays, J., Perona, P., Ramanan, D.,
  Doll{\'a}r, P., Zitnick, C.L.: Microsoft coco: Common objects in context. In:
  The European Conference on Computer Vision (ECCV) (2014)

\bibitem{lu2019vilbert}
Lu, J., Batra, D., Parikh, D., Lee, S.: Vilbert: Pretraining task-agnostic
  visiolinguistic representations for vision-and-language tasks.
  arXiv:1908.02265  (2019)

\bibitem{MACKIEWICZ1993303}
Maćkiewicz, A., Ratajczak, W.: Principal components analysis (pca). Computers
  and Geosciences  \textbf{19}(3),  303 -- 342 (1993)

\bibitem{miyazaki2016cross}
Miyazaki, T., Shimizu, N.: Cross-lingual image caption generation. In:
  Conference of the Association for Computational Linguistics (ACL) (2016)

\bibitem{Nguyen_2019_CVPR}
Nguyen, D.K., Okatani, T.: Multi-task learning of hierarchical vision-language
  representation. In: The IEEE Conference on Computer Vision and Pattern
  Recognition (CVPR) (2019)

\bibitem{pires2019multilingual}
Pires, T., Schlinger, E., Garrette, D.: How multilingual is multilingual bert?
  arXiv:1906.01502  (2019)

\bibitem{plummer2015flickr30k}
Plummer, B.A., Wang, L., Cervantes, C.M., Caicedo, J.C., Hockenmaier, J.,
  Lazebnik, S.: Flickr30k entities: Collecting region-to-phrase correspondences
  for richer image-to-sentence models. In: The IEEE International Conference on
  Computer Vision (ICCV) (2015)

\bibitem{fasterrcnn}
Ren, S., He, K., Girshick, R., Sun, J.: Faster r-cnn: Towards real-time object
  detection with region proposal networks. In: Advances in Neural Information
  Processing Systems (NeurIPS) (2015)

\bibitem{bilingual1}
Smith, S.L., Turban, D.H.P., Hamblin, S., Hammerla, N.Y.: Offline bilingual
  word vectors, orthogonal transformations and the inverted softmax.
  arXiv:1702.03859  (2017)

\bibitem{su2019vlbert}
Su, W., Zhu, X., Cao, Y., Li, B., Lu, L., Wei, F., Dai, J.: Vl-bert:
  Pre-training of generic visual-linguistic representations. arXiv:1908.08530
  (2019)

\bibitem{tan2019lxmert}
Tan, H., Bansal, M.: Lxmert: Learning cross-modality encoder representations
  from transformers. In: Proceedings of the 2019 Conference on Empirical
  Methods in Natural Language Processing (EMNLP) (2019)

\bibitem{vaswani_transformer}
Vaswani, A., Shazeer, N., Parmar, N., Uszkoreit, J., Jones, L., Gomez, A.N.,
  Kaiser, L.u., Polosukhin, I.: Attention is all you need. In: Guyon, I.,
  Luxburg, U.V., Bengio, S., Wallach, H., Fergus, R., Vishwanathan, S.,
  Garnett, R. (eds.) Advances in Neural Information Processing Systems
  (NeurIPS), pp. 5998--6008 (2017)

\bibitem{wang2018learning}
Wang, L., Li, Y., Huang, J., Lazebnik, S.: Learning two-branch neural networks
  for image-text matching tasks. IEEE Transactions on Pattern Analysis and
  Machine Intelligence (TPAMI)  \textbf{41}(2),  394--407 (2018)

\bibitem{Wehrmann_2019_ICCV}
Wehrmann, J., Souza, D.M., Lopes, M.A., Barros, R.C.: Language-agnostic
  visual-semantic embeddings. In: The IEEE International Conference on Computer
  Vision (ICCV) (2019)

\bibitem{wu2019beto}
Wu, S., Dredze, M.: Beto, bentz, becas: The surprising cross-lingual
  effectiveness of bert. arXiv:1904.09077  (2019)

\bibitem{young2014image}
Young, P., Lai, A., Hodosh, M., Hockenmaier, J.: From image descriptions to
  visual denotations: New similarity metrics for semantic inference over event
  descriptions. Transactions of the Association for Computational Linguistics
  (TACL)  \textbf{2},  67--78 (2014)

\bibitem{VCR}
Zellers, R., Bisk, Y., Farhadi, A., Choi, Y.: From recognition to cognition:
  Visual commonsense reasoning. In: The IEEE Conference on Computer Vision and
  Pattern Recognition (CVPR) (2019)

\end{thebibliography}
\newpage

\section{Supplementary Material}
\subsection{Data Augmentation}
We augment the multilingual datasets MSCOCO~\cite{li2019coco,lin2014microsoft,miyazaki2016cross} and Multi30K~\cite{barrault2018findings,elliott2017findings,elliott2016multi30k} with translations from languages with human-generated annotations to other languages using Google Translate.

Tables~\ref{tab:mt_coco_breakdown} and~\ref{tab:mt_m30k_breakdown} show what translations were performed for MSCOCO and Multi30K, respectively. The column X refers to all other languages that consist entirely of translations to create the total set of ten languages; \ie for MSCOCO, $X \in $ German, French, Czech, Arabic, Afrikaans, Korean, Russian, and for Multi30K, $X \in$ Chinese, Japanese, Arabic, Afrikaans, Korean, Russian. We compare the effect of using human-generated vs. machine translated sentences at test time in Section~\ref{subsec:trans_compare}.

\begin{table}[h]
    \centering
    \caption{Dataset Augmentation for MSCOCO. Arrows signify the use of machine translation, and X refers to all other languages in the total set of ten}
    \label{tab:mt_coco_breakdown}
  \begin{tabular}{|c|c|c|c|c|}
    \hline
    Annotation Type & En & Cn & Ja & X\\
    \hline
     Human Generated & MSCOCO~\cite{lin2014microsoft} & COCO-CN\cite{li2019coco}& YJ Captions~\cite{miyazaki2016cross}& --\\
     \hline
 \multirow{3}{*}{Translations} & Cn $\rightarrow$ En & En $\rightarrow$ Cn & En $\rightarrow$ Ja & En $\rightarrow$ X\\
  & Ja $\rightarrow$ En & &   & \\
  \hline
\end{tabular}
\end{table}

\begin{table}[h]
    \centering
    \caption{Dataset Augmentation for Multi30K. Arrows signify the use of machine translation, and X refers to all other languages in the total set of ten}
    \label{tab:mt_m30k_breakdown}
  \begin{tabular}{|c|c|c|c|c|c|}
    \hline
    Annotation Type & En & De & Fr & Cs & X\\
    \hline
     Human Generated & Flickr30K~\cite{young2014image} & Multi30K~\cite{elliott2016multi30k}& Multi30K~\cite{elliott2017findings} & Multi30K~\cite{barrault2018findings} & --\\
     \hline
 \multirow{3}{*}{Translations} & De $\rightarrow$ En & En $\rightarrow$ De & En $\rightarrow$ Fr & En $\rightarrow$ Cs & En $\rightarrow$ X\\
  & Fr $\rightarrow$ En && &&\\
  & Cs $\rightarrow$ En &&& &\\
  \hline
\end{tabular}
\end{table}
\subsection{Model Parameters}
\subsubsection{Exploration Parameters}
One component of SMALR is the Hybrid Embedding Model (HEM), which makes use of both language-specific and language-agnostic representations. The Language-Agnostic (LA) baseline refers to only using the shared latent vocabulary, which consists of 40K tokens. We found experimentally that using exploration parameters $p = 0.2$ and $M = 20$ improves downstream performance when using the latent vocabulary.  These exploration parameters are used to force the model to randomly select from a set of similar tokens during training rather than always choosing the best matched token in the language-agnostic vocabulary (described in Section 3.1 of the main paper).  Tables~\ref{tab:coco_explore} and~\ref{tab:m30k_explore} demonstrate the difference in mean Recall for image-sentence retrieval with and without our exploration parameters. 

Since we find that using the exploration parameters when learning the mapping to the latent vocabulary improves performance, we use them for both the Language-Agnostic and HEM results (and thus is included in the final SMALR training paradigm).

\begin{table}[h]
\setlength{\tabcolsep}{1.pt}
\begin{center}
\caption{MSCOCO Language-Agnostic (LA) Ablation}
\label{tab:coco_explore}
\begin{tabular}{|l|c|c|c|c|c|c|c|c|c|c|c|c|}
\hline
Model & En & De\footnotemark[1] & Fr\footnotemark[1] & Cs\footnotemark[1] & Cn & Ja & Ar\footnotemark[1] & Af\footnotemark[1] & Ko\footnotemark[1] & Ru\footnotemark[1] & HA & A\\
\hline
\hline
LA & 64.2 & 58.8 & 58.3 & 52.1 & 59.0 & 63.2 & 61.9 & 65.3 & 58.6 & 58.5 & 58.3 & 60.0\\
LA + Explore & \textbf{65.5} & \textbf{61.3} & \textbf{59.9}& \textbf{54.0} & \textbf{59.4} & \textbf{64.7} & \textbf{63.9} & \textbf{66.5} & \textbf{60.3} & \textbf{60.3} & \textbf{60.2} & \textbf{61.6} \\
\hline
\end{tabular}
\end{center}
\footnotemark[1]uses translations from English for testing\\ 
\end{table}

\begin{table}[h]
\setlength{\tabcolsep}{1.pt}
\begin{center}
\caption{Multi30K Language-Agnostic (LA) Ablation}
\label{tab:m30k_explore}
\begin{tabular}{|l|c|c|c|c|c|c|c|c|c|c|c|c|}
\hline
Model & En & De & Fr & Cs & Cn\footnotemark[1] & Ja\footnotemark[1] & Ar\footnotemark[1] & Af\footnotemark[1] & Ko\footnotemark[1] & Ru\footnotemark[1] & HA & A\\
\hline
\hline
LA & 73.9 & 73.0 & 71.7 & 72.9 & 72.0 & 70.8 & 72.8 & 72.0 & 69.7 & 72.0 & 72.2 & 72.1\\
LA + Explore & \textbf{75.0} & \textbf{74.3} & \textbf{74.1} & \textbf{73.4} & \textbf{72.3} & \textbf{72.1} & \textbf{74.4} & \textbf{74.7} & \textbf{71.6} & \textbf{72.7} & \textbf{73.1} & \textbf{73.5}\\
\hline
\end{tabular}
\end{center}
\footnotemark[1]uses translations from English for testing\\ 
\end{table}
\clearpage
\subsubsection{Loss Parameters}
\begin{figure}
    \centering
    \includegraphics[scale=0.6]{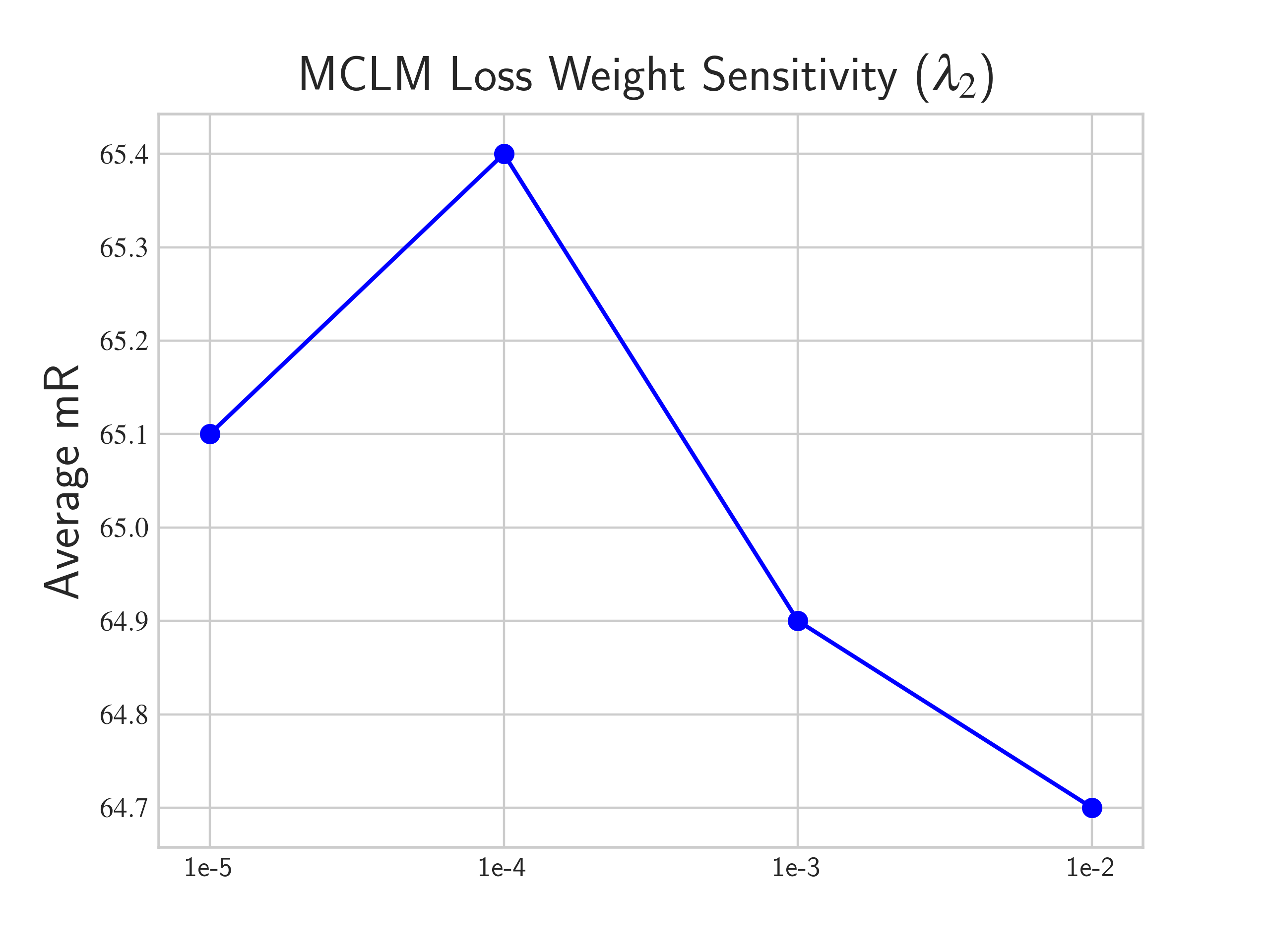}
    \caption{M30K ablation results for the $\lambda_2$ parameter. As shown in Eq. 4 of the main paper, $\lambda_2$ determines the contribution of the masking loss to the total SMALR training loss}
    \label{fig:m30k_loss}
\end{figure}

Training SMALR did not require significant hyperparameter tuning. We found the results were not sensitive to our choice of lambdas used in the SMALR loss, as defined in Eq. 4 of the main paper. Therefore, $\lambda_1$, $\lambda_3$ and $\lambda_4$ were kept the same as in prior work [16] for consistency. The parameter $\lambda_2$ is associated with the MCLM masking loss we introduce, which is determined by grid search over powers of ten on the validation set. On Multi30K, the average mR for SMALR when varying $\lambda_2$ has a performance range under one point, see Figure~\ref{fig:m30k_loss} below for exact values.

\subsection{Qualitative Results}
We provide two examples for both MSCOCO and Multi30K which show the effect of the Cross-Lingual Consistency (CLC) module used with SMALR. We report results for the CLC-C variant, which makes use of a simple MLP classifier to aggregate scores across language. For a given text query, if it is human generated, we translate it to all other languages and use the predictions from these translations as input to our CLC-C module.

\begin{figure}[ht]
    \centering
    \includegraphics[scale=0.3]{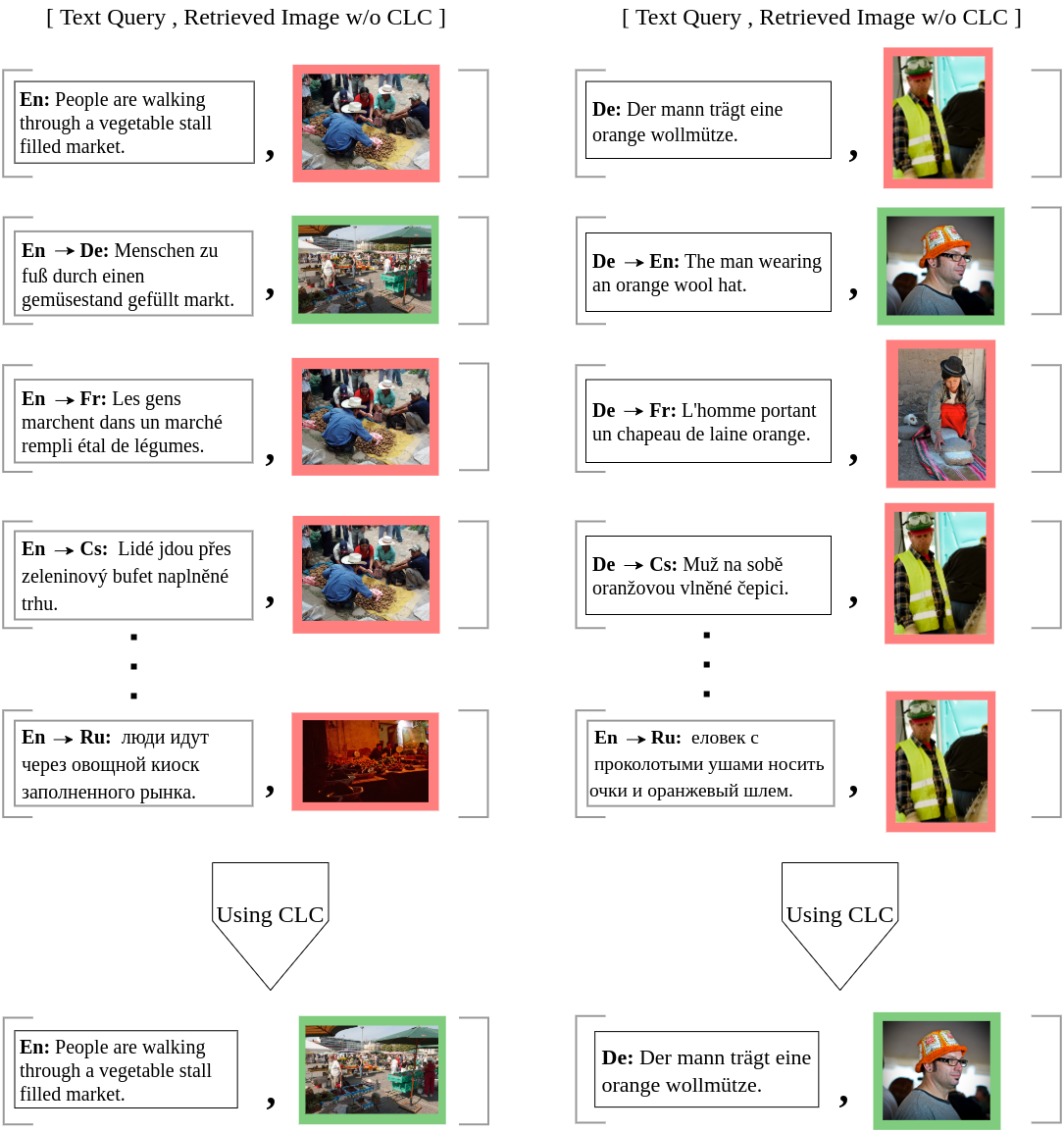}
    \caption{Example of the benefits of using the CLC module on Multi30K}
    \label{fig:m30k_qualitative}
\end{figure}

On the left hand side of Figure~\ref{fig:m30k_qualitative}, the original text query is in English and its matching image is incorrectly retrieved, as shown by the red bounding box. However, when CLC-C is used, SMALR is able to correctly retrieve the matching image, as a subset of the translated sentences do correctly retrieve the ground truth image (\eg  the German translation). On the right hand side of Figure~\ref{fig:m30k_qualitative}, we also see the same benefit for an original text query in German which is aided by English translations. These two examples demonstrate the benefit of CLC-C for R@1, as CLC-C now correctly retrieves the ground truth image. Additionally, these samples show that every language does not have to make the correct prediction; the CLC-C module can learn to combine predictions to improve performance. As we can see in Figure~\ref{fig:m30k_qualitative}, the images incorrectly retrieved for the original English and German queries ``People are walking through a vegetable stall filled market'' and ``Der mann trägt eine orange wollmütze'' contain very similar objects and colors to their respective ground truth images, but these errors are remedied when considering all languages.

\begin{figure}[ht]
    \centering
    \includegraphics[scale=0.3]{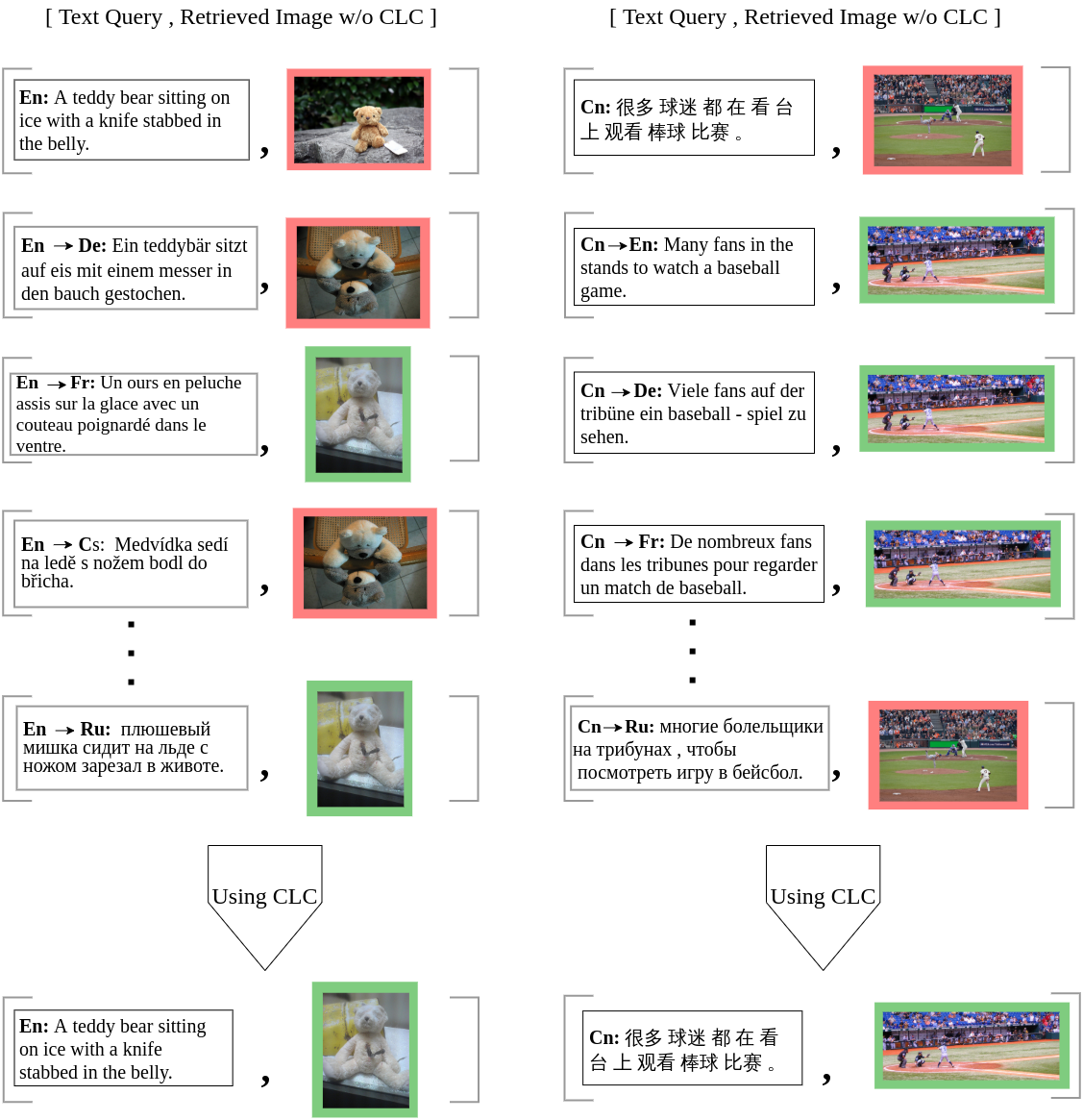}
    \caption{Example of the benefits of using the CLC module on MSCOCO}
    \label{fig:coco_qualitative}
\end{figure}

In Figure~\ref{fig:coco_qualitative}, there are two examples for MSCOCO, with original text queries in English and Chinese. Both examples have many translated queries which are able to correctly retrieve the ground truth image, such as French and Russian for English, and English, German, and French (among others) for Chinese.
We see again that the original incorrectly retrieved image contains very similar visual semantics (\eg teddy bear for English, baseball field for Chinese) to the ground truth, and the translated sentences help disambiguate subtle details.

\subsection{Masked Cross-Language Modeling Example}
\begin{figure}
    \centering
    \includegraphics[scale=0.18]{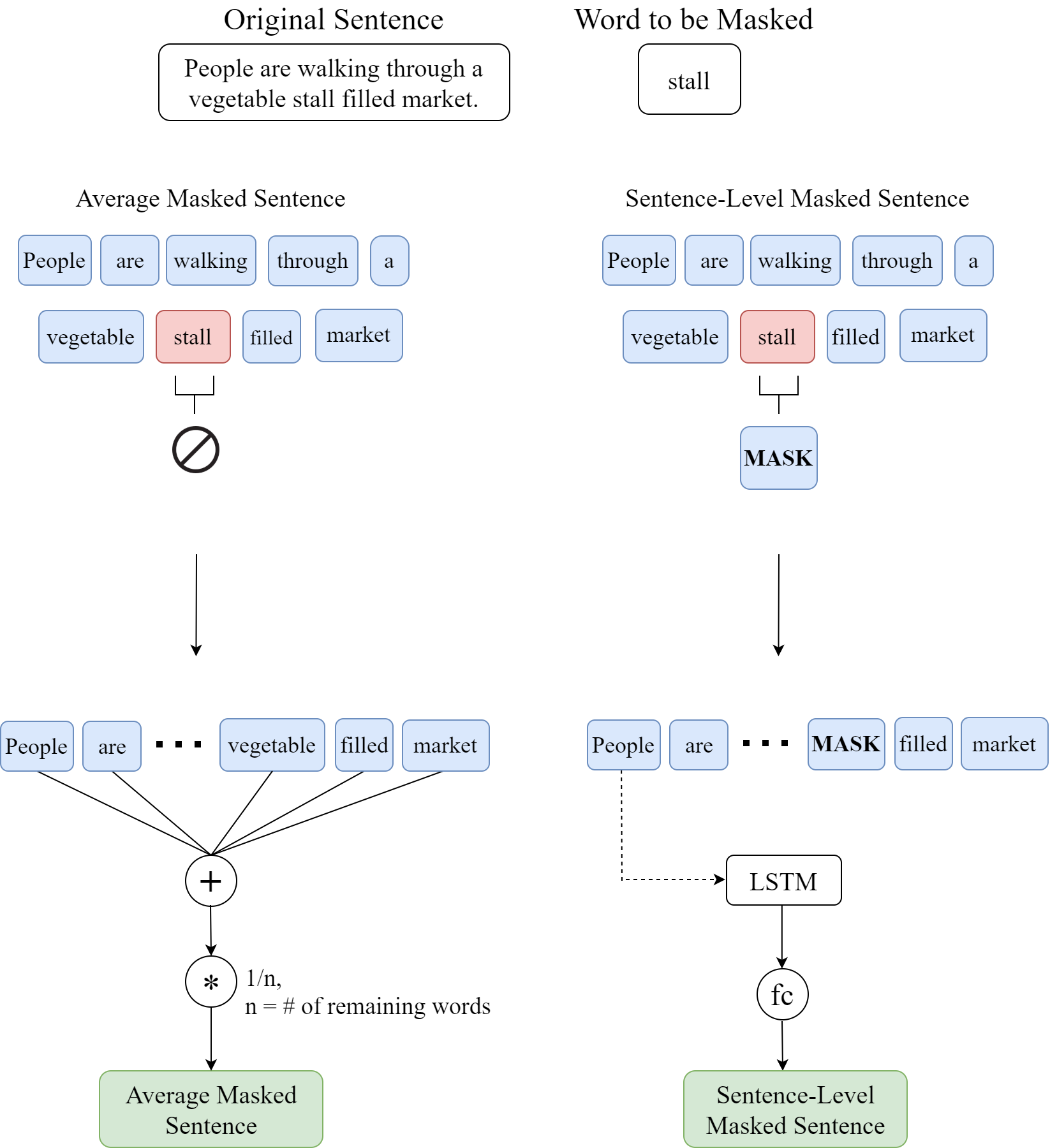}
    \caption{Variants of masking used in the MCLM module}
    \label{fig:mclm}
\end{figure}

SMALR's Masked Cross-Language Model (MCLM) uses two language representations to compute its total loss, namely an average representation, and a sentence-level LSTM representation. The average masked sentence simply removes masked words and then averages each word embedding over the shorter version of the original sentence before predicting the masked token. The masked sentence-level representation retains the same number of words by replacing the masked words with a special [MASK] token; not only does this retain the total word count for a given query, it also maintains grammatical structure by using a LSTM. This representation is passed through a LSTM and fully connected layer before being used to predict the masked token. Figure~\ref{fig:mclm} provides an example of this process; the word boxes represent word embeddings.  See Section 3.2 of the main paper for a description of how these representations are used.
\subsection{Extended Image-Sentence Retrieval Results}

We provide all recall values (Recall@K for K$\in \{1,5,10\}$) for all ten languages on image-sentence retrieval with MSCOCO and Multi30K. I-to-S signifies the image to sentence retrieval direction, and S-to-I the sentence to image direction. We shorten ``Language-Agnostic'' to ``LA'' and CLC-A, CLC-C to A and C, respectively, due to space constraints. Lastly, ``Prior'' refers to prior work, ``Adapted'' refers to prior work that has been adapted to our testing scenario using the author's publicly available code, and ``Ours'' refers to our SMALR model variants. The number preceding a model refers to the number of languages it was trained on, \eg (3-4) MULE signifies MULE~\cite{kimMULEAAAI2020} trained on three languages (English, Chinese, Japanese) on MSCOCO, and four on Multi30K (English, German, French, Czech).

\begin{table}[ht]
\setlength{\tabcolsep}{1.pt}
\begin{center}
\caption{English bidirectional image-sentence retrieval results using human-generated sentences}
\label{tab:mscoco_en}
\begin{tabular}{|l|c|c|c|c|c|c|c|c|c|c|c|c|c|c|}
\hline
\multirow{3}{*}{Model} & \multicolumn{7}{c|}{MSCOCO} & \multicolumn{7}{c|}{Multi30K} \\
\cline{2-8}
\cline{9-15}
 & \multicolumn{3}{c|}{I-to-S} & \multicolumn{3}{c|}{S-to-I} & \multirow{2}{*}{mR} & \multicolumn{3}{c|}{I-to-S} & \multicolumn{3}{c|}{S-to-I} & \multirow{2}{*}{mR} \\
\cline{2-7}
\cline{9-14}
& r@1 & r@5 & r@10 & r@1 & r@5 & r@10& & r@1 & r@5 & r@10 & r@1 & r@5 & r@10& \\
\hline
\hline
\textbf{(a)} Prior & & & & & & & & & & & & & &\\
Trans. to En&  58.6 & 86.5& 94.1 & 45.5& 79.6 & 89.5 & 75.6  & 58.3 & 82.9 & 90.4& 41.7 & 72.0 & 81.2 & 71.1 \\
EmbN &61.8 & 87.6 & 94.1 & 47.5 & 79.8 & 89.8 & 76.8 & 57.9 & 84.5 & 90.9 & 44.3& 72.7& 84.7 & 72.0\\
PAR. EmbN &63.1 &89.1 &94.1 &49.2& 82.5 &91.5 &78.3 &52.4& 80.1& 87.7 &41.6 &71.5& 80.7& 69.0 \\
(3-4) MULE &63.9& 90.2 &95.8 &50.9 &83.5& 92.4 &79.5& 54.2 & 82.0 & 89.9 & 41.9& 72.5 & 81.1 & 70.3
 \\
\hline
\hline
\textbf{(b)} Adapted & & && & & & & && & & & & \\
(1) S-LIWE & \textbf{66.8}&91.2& 96.6&52.4&85.1&93.5&80.9
&  \textbf{65.5}& 88.9 & \textbf{95.1} & 46.9 & 77.2 & 84.5 & \textbf{76.3}
\\
 (2) S-LIWE & 62.3& 87.3 & 94.6 & 48.3& 80.7&91.0& 77.4
& 64.5&88.1&94.3&46.4&75.8&84.5&75.6
\\
(10) S-LIWE & 61.8&88.2&94.8&47.9&80.3&90.5& 77.3
& 63.8 & 88.0 & 93.4 & 46.4 & 75.4 & 84.3 & 75.2
 \\
(10) L-LIWE&63.8&90.2&95.6&50.1&82.9&92.2&79.1
 & 63.9 & \textbf{89.0} & 94.2 & 46.9 & 76.8 & 84.8 & 75.9
\\
(10) MULE &63.8& 88.9& 95.5 & 50.5 & 83.2 & 92.0 & 79.0& 
 55.2 & 82.1 & 90.7 & 42.2 & 72.2 & 81.8 & 70.7 \\
\hline
\hline
\textbf{(c)} Ours & & & & & & & & & & & & & &\\ 
LA & 56.4 & 84.9 & 92.3 & 46.0 & 80.5 & 90.2 & 75.0 & 48.1 & 77.2 & 86.9 & 36.5& 67.1& 77.2& 65.5\\
HEM & 61.6 & 89.0 & 95.4 & 50.5 & 83.3 & 92.4 & 78.7  & 51.3 & 79.9 & 88.4 & 41.8 & 72.1 & 81.5 & 69.2\\
SMALR & 62.9 & 89.2 & 95.8 & 51.1 & 84.0 & 92.5 & 79.3  &  52.0 & 81.1& 88.4& 41.8 & 72.4& 82.1 & 69.6\\
SMALR-A& 66.6 & 91.1& 97.3 & 52.8 & 85.7 & 93.4 & 81.2  & 59.4 & 83.7 & 90.2 & 47.5 & 77.5 & 86.1 & 74.1\\
SMALR-C& 66.5 & \textbf{91.3}& \textbf{97.5} & \textbf{53.6} & \textbf{86.2} & \textbf{94.0} & \textbf{81.5} &60.2 & 83.8 & 91.0 & \textbf{47.9} & \textbf{77.9} & \textbf{86.3} & 74.5\\
\hline
\end{tabular}
\end{center}
\end{table}

\begin{table}[ht]
\setlength{\tabcolsep}{1.pt}
\begin{center}
\caption{German bidirectional image-sentence retrieval results using sentences translated from English into German for testing on MSCOCO and human-generated sentences on Multi30K}
\label{tab:mscoco_de}
\begin{tabular}{|l|c|c|c|c|c|c|c|c|c|c|c|c|c|c|}
\hline
\multirow{3}{*}{Model} & \multicolumn{7}{c|}{MSCOCO} & \multicolumn{7}{c|}{Multi30K} \\
\cline{2-8}
\cline{9-15}
 & \multicolumn{3}{c|}{I-to-S} & \multicolumn{3}{c|}{S-to-I} & \multirow{2}{*}{mR} & \multicolumn{3}{c|}{I-to-S} & \multicolumn{3}{c|}{S-to-I} & \multirow{2}{*}{mR} \\
\cline{2-7}
\cline{9-14}
& r@1 & r@5 & r@10 & r@1 & r@5 & r@10& & r@1 & r@5 & r@10 & r@1 & r@5 & r@10& \\
\hline
\hline
\textbf{(a)} Prior& & & & & & & & & & & & & &\\
Trans. To En & -- & -- & -- & -- & -- & -- & -- & 34.1 & 60.4 & 71.1 & 19.6& 47.4& 58.5 &48.5\\
 EmbN & -- & -- & -- & -- & -- & -- & -- & 46.6 & 73.9 & 82.2 & 31.3 & 59.1 & 69.0& 60.3\\
 PAR. EmbN& -- & -- & -- & -- & -- & -- & --  & 46.1 &76.3& 83.2& 34.4 &62.5 &73.0& 62.6 \\
 (3-4) MULE& -- & -- & -- & -- & -- & -- & --& 49.7 & 77.7 & 85.7 & 34.6 & 63.4 & 73.5 & 64.1\\
\hline
\hline
\textbf{(b)} Adapted& & & & & & & & & & & & & &\\
(1) S-LIWE& -- & -- & -- & -- & -- & -- & -- &\textbf{61.1}&\textbf{86.6}&\textbf{92.7}&42.0&69.9&80.0&\textbf{72.1}
 \\
 (2) S-LIWE& -- & -- & -- & -- & -- & -- & -- &51.2&80.2&88.4&35.7&65.7&75.2&66.1
\\
 (10) S-LIWE& 49.8&79.1&87.3&36.6&69.4&82.4&67.4
 & 50.5 & 79.0 & 88.0 & 34.9 & 64.3 & 74.3 & 65.2
\\
 (10) L-LIWE&  52.1 & 84.9 & 92.6 & 39.3 & 73.4 & 85.0&71.2
 & 51.1 & 80.5 & 89.8 & 35.9 & 66.6 & 76.1 & 66.7
\\
(10) MULE&59.1 &88.7 & 94.9 & 48.5 & 81.3 & 90.6 & 77.2& 45.8 & 75.8 & 85.2 & 35.1 & 64.6 & 75.3 & 63.6 \\
\hline
\hline
\textbf{(c)} Ours&&&&&&&& & & & & & &\\
LA & 54.4& 86.2& 93.1 & 44.5& 78.6& 88.7 & 74.3 & 44.0 & 75.4 & 85.1 & 32.2& 59.7& 71.0& 61.3  \\
HEM & 59.2 & 87.2 & 95.1 & 49.1 & 81.8 & 91.4 & 77.3 & 49.2 & 75.4& 83.2 & 34.5& 62.0 & 72.4 & 62.8\\
SMALR & \textbf{61.2} & \textbf{89.2} & \textbf{96.2} & \textbf{49.6} & \textbf{82.3} & \textbf{91.8} & \textbf{78.4}  &  49.9& 75.8& 85.0 & 36.9 & 65.4 & 75.4 & 64.7\\
SMALR-A & -- & -- & -- & -- & -- & -- & -- & 53.0 &77.6 & 85.8& 41.9& 72.9 & 82.3 & 68.9 \\
SMALR-C & -- & -- & -- & -- & -- & -- & -- &52.9 & 78.8 & 87.0 & \textbf{42.6}& \textbf{74.2} & \textbf{83.1} & 69.8\\
\hline
\end{tabular}
\end{center}
\end{table}

\begin{table}[ht]
\setlength{\tabcolsep}{1.pt}
\begin{center}
\caption{French bidirectional image-sentence retrieval results using sentences translated from English into French for testing on MSCOCO and human-generated sentences on Multi30K}
\label{tab:mscoco_fr}
\begin{tabular}{|l|c|c|c|c|c|c|c|c|c|c|c|c|c|c|}
\hline
\multirow{3}{*}{Model} & \multicolumn{7}{c|}{MSCOCO} & \multicolumn{7}{c|}{Multi30K} \\
\cline{2-8}
\cline{9-15}
 & \multicolumn{3}{c|}{I-to-S} & \multicolumn{3}{c|}{S-to-I} & \multirow{2}{*}{mR} & \multicolumn{3}{c|}{I-to-S} & \multicolumn{3}{c|}{S-to-I} & \multirow{2}{*}{mR} \\
\cline{2-7}
\cline{9-14}
& r@1 & r@5 & r@10 & r@1 & r@5 & r@10& & r@1 & r@5 & r@10 & r@1 & r@5 & r@10& \\
\hline
\hline
\textbf{(a)} Prior& & & & & & & & & & & & & &\\
Trans. to En& -- & -- & -- & -- & -- & -- & --&22.5& 52.5& 63.0 & 25.1 & 53.1 & 63.9 & 46.7\\
 EmbN& -- & -- & -- & -- & -- & -- & --& 31.0 &60.4& 71.0 & 35.2 & 60.3 & 70.8& 54.8\\
 PAR. EmbN& -- & -- & -- & -- & -- & -- & -- &37.6 &66.0 &77.4 &37.8 &66.4 &78.2 &60.6  \\
 (3-4) MULE& -- & -- & -- & -- & -- & -- & --&38.0 &68.4 & 80.0 & 38.2 & 68.9 & 80.3 & 62.3   \\
\hline
\hline
\textbf{(b)} Adapted& & & & & & & & & & & & & &\\
 (10) S-LIWE& 50.8& 79.3&90.4&36.5&70.7&83.2&68.5
& 39.0 & 39.0 & 51.6 & 37.3 & 66.7 & 77.3 & 51.8
  \\
 (10) L-LIWE& 51.8&81.3&92.2&39.0&73.1&84.7&70.3
 & 40.6 & 40.7 & 54.7 & 37.8 & 69.3 & 78.1 & 53.5
  \\
(10) MULE & \textbf{60.3 }& 86.9 & 94.3 & 47.8 & 81.3 & 90.4 & 76.8 &39.2 &70.9 & 80.7 & 38.8 & 70.5 & 80.2 & 63.4
 \\
\hline
\hline
\textbf{(c)} Ours & & & & & & & & & & & & & &\\
LA& 54.8 & 83.6 & 92.6 & 44.8 & 79.4 & 89.7 & 74.1 &35.1 & 65.8 & 76.0 & 39.5 & 65.6 & 77.2 & 59.9 \\
HEM & 57.6 & 87.0 & 94.0 & 48.0 & 80.7 & \textbf{91.1} & 76.4& 38.1& 70.5 &80.6 & 40.2 & 69.5 & 80.6 &63.3 \\
SMALR & 59.6 & \textbf{89.7} & \textbf{95.9} & \textbf{48.7} & \textbf{81.9} & 91.0 & \textbf{77.8} & 40.6 & 70.7 & 81.8 & 41.1 & 71.8 & 80.7 & 64.5\\
SMALR-A & -- & -- & -- & -- & -- & -- & --& 40.3 & \textbf{73.4} & 80.9 & 42.2 & 72.8 & 81.8& 65.2\\
SMALR-C & -- & -- & -- & -- & -- & -- & --&\textbf{41.1}&\textbf{73.4}& \textbf{82.5} & \textbf{42.6} & \textbf{73.0} & \textbf{82.9} & \textbf{65.9}\\
\hline
\end{tabular}
\end{center}
\end{table}

\begin{table}[ht]
\setlength{\tabcolsep}{1.pt}
\begin{center}
\caption{Czech bidirectional image-sentence retrieval results using sentences translated from English into Czech for testing on MSCOCO and human-generated sentences on Multi30K}
\label{tab:mscoco_cs}
\begin{tabular}{|l|c|c|c|c|c|c|c|c|c|c|c|c|c|c|}
\hline
\multirow{3}{*}{Model} & \multicolumn{7}{c|}{MSCOCO} & \multicolumn{7}{c|}{Multi30K} \\
\cline{2-8}
\cline{9-15}
 & \multicolumn{3}{c|}{I-to-S} & \multicolumn{3}{c|}{S-to-I} & \multirow{2}{*}{mR} & \multicolumn{3}{c|}{I-to-S} & \multicolumn{3}{c|}{S-to-I} & \multirow{2}{*}{mR} \\
\cline{2-7}
\cline{9-14}
& r@1 & r@5 & r@10 & r@1 & r@5 & r@10& & r@1 & r@5 & r@10 & r@1 & r@5 & r@10& \\
\hline
\hline
\textbf{(a)} Prior& & & & & & & & & & & & & &\\
Trans. to En& -- & -- & -- & -- & -- & -- & --& 23.0 & 50.9 & 64.7 & 25.1 & 53.4 & 64.2 & 46.9\\
EmbN& -- & -- & -- & -- & -- & -- & -- & 26.2 &51.3 & 62.5 & 26.8 & 50.3 & 60.8& 46.3  \\
PAR. EmbN& -- & -- & -- & -- & -- & -- & -- &31.4 &  58.2& 70.1 &33.1 & 60.4 & 71.6& 54.1 \\
(3-4) MULE& -- & -- & -- & -- & -- & -- & -- & 34.3 & 63.2 & 74.2 & 35.3 & 63.6 & 75.5 & 57.7 \\
\hline
\hline
\textbf{(b)} Adapted& & & & & & & & & & & & & &\\
(10) S-LIWE&  46.8&79.8&90.3&34.6&68.2&82.0& 66.9
& 36.5 & 36.5 & 50.0 & 37.6 & 64.3 & 75.2 & 50.0
\\
(10) L-LIWE& 50.7 & 82.3 & 92.1 & 37.6 & 72.8 & 84.8&70.1& 37.6 & 37.6 & 52.9 & 38.1 & 66.2 & 75.2 & 51.3
 \\
(10) MULE& 61.6 & 88.7 & 94.8 & 48.8 & 81.5 & 91.1 & 77.8 & 37.0 & 66.3 & 76.4 & 37.5 & 64.6 & 74.8 & 59.4\\
\hline
\hline
\textbf{(c)} Ours& & & & & & & & & & & & & &\\
LA &55.3 & 84.6 & 92.4 & 43.5 & 76.9 & 87.8 &73.4 & 31.0 & 59.6 & 71.1 & 32.5 & 58.5 & 71.5 & 54.0\\
HEM & 59.9 & 88.4 & 95.4 & \textbf{49.2} & \textbf{82.5} & \textbf{91.7} & 77.9& 35.0 & 66.9 & 77.4 & 36.1 & 67.4 & 77.2& 60.0\\
SMALR & \textbf{63.2} & \textbf{89.6} & \textbf{95.7} & \textbf{49.2} & 82.4 & 91.6 & \textbf{78.6}& 36.5 &69.0 & 78.0 & 36.7 & 68.0 & 78.2 & 61.1\\
SMALR-A & -- & -- & -- & -- & -- & -- & --& 41.1 & \textbf{70.7} & 80.4 & 39.9 & \textbf{71.8} & \textbf{83.0} & 64.5\\
SMALR-C & -- & -- & -- & -- & -- & -- & --& \textbf{41.9} & \textbf{70.7}& \textbf{81.1}& \textbf{40.5}& 71.7 & 82.8 & \textbf{64.8}\\
\hline
\end{tabular}
\end{center}
\end{table}

\begin{table}[ht]
\setlength{\tabcolsep}{1.pt}
\begin{center}
\caption{Chinese bidirectional image-sentence retrieval results using sentences translated from English into Chinese for testing on Multi30K and human-generated sentences on MSCOCO}
\label{tab:mscoco_cn}
\begin{tabular}{|l|c|c|c|c|c|c|c|c|c|c|c|c|c|c|}
\hline
\multirow{3}{*}{Model} & \multicolumn{7}{c|}{MSCOCO} & \multicolumn{7}{c|}{Multi30K} \\
\cline{2-8}
\cline{9-15}
 & \multicolumn{3}{c|}{I-to-S} & \multicolumn{3}{c|}{S-to-I} & \multirow{2}{*}{mR} & \multicolumn{3}{c|}{I-to-S} & \multicolumn{3}{c|}{S-to-I} & \multirow{2}{*}{mR} \\
\cline{2-7}
\cline{9-14}
& r@1 & r@5 & r@10 & r@1 & r@5 & r@10& & r@1 & r@5 & r@10 & r@1 & r@5 & r@10& \\
\hline
\hline
\textbf{(a)} Prior& & & & & & & & & & & & & &\\
Trans. to En & 45.9 & 79.8 & 89.2 & 47.8 & 81.1 & 89.4 & 72.2&-- & -- & -- & -- & -- & -- & -- \\
EmbN& 49.6 &81.6 &90.0 &47.8 &82.1& 90.0 &73.5 & -- & -- & -- & -- & -- & -- & -- \\
PAR. EmbN& 47.9 &81.4 &91.1 &47.5 & 81.6 & 91.2 &73.5 & -- & -- & -- & -- & -- & -- & -- \\
(3-4) MULE& 51.1 & 82.6 & 91.6 & 49.1& 82.4 & 91.9 & 74.8 & -- & -- & -- & -- & -- & -- & --  \\
\hline
\hline
\textbf{(b)} Adapted& & & & & & & & & & & & & &\\
(10) S-LIWE& 45.1&76.4&88.1&32.7&66.0&79.6&64.5
 & 39.3 & 68.1 & 79.2 & 24.2 & 51.0 & 62.5 & 54.1
 \\
(10) L-LIWE&  51.4&82.6&91.3&38.1&72.2&84.6&70.0
& 42.6 & 72.4 & 82.4 & 26.0 & 53.6 & 64.7 & 56.9
 \\
(10) MULE& 50.8 & 84.0 & 92.5 & 50.3 & 83.6 & 92.4 & 75.6 & \textbf{47.4} & \textbf{77.0} & \textbf{85.8} & 35.4 & 64.9 & 74.4 & \textbf{64.2}
  \\
\hline
\hline
\textbf{(c)} Ours& & & & & & & & & & & & & &\\
LA& 46.0 & 79.6& 90.7 & 45.9 & 80.6 & 91.1& 72.3&42.2 & 72.0 & 81.6 & 30.6 & 59.8 & 70.0 & 59.4\\
HEM & 53.2& 85.0 & 93.2 & 51.3 & 84.6 & 93.0 & 76.7& 44.1 & 74.7 & 84.4 & 33.8 & 63.3 & 74.4& 62.4\\
SMALR & 51.2  & 86.5 & 93.8 & 50.6 & 84.7 & 93.3 & 76.7& 45.8 &\textbf{77.0}& 85.0 & \textbf{35.8} & \textbf{65.1} & \textbf{75.5} & 64.0\\
SMALR-A & 57.5 & 87.3 & 94.9 & 54.8 & 87.7 & 95.2 & 79.6& -- & -- & -- & -- & -- & -- & --\\
SMALR-C & \textbf{58.0} & \textbf{87.8} & \textbf{95.4} & \textbf{55.3} & \textbf{88.2} & \textbf{95.7} & \textbf{80.1}& -- & -- & -- & -- & -- & -- & --\\
\hline
\end{tabular}
\end{center}
\end{table}

\begin{table}[ht]
\setlength{\tabcolsep}{1.pt}
\begin{center}
\caption{Japanese bidirectional image-sentence retrieval results using sentences translated from English into Japanese for testing on Multi30K and human-generated sentences on MSCOCO}
\label{tab:mscoco_jp}
\begin{tabular}{|l|c|c|c|c|c|c|c|c|c|c|c|c|c|c|}
\hline
\multirow{3}{*}{Model} & \multicolumn{7}{c|}{MSCOCO} & \multicolumn{7}{c|}{Multi30K} \\
\cline{2-8}
\cline{9-15}
 & \multicolumn{3}{c|}{I-to-S} & \multicolumn{3}{c|}{S-to-I} & \multirow{2}{*}{mR} & \multicolumn{3}{c|}{I-to-S} & \multicolumn{3}{c|}{S-to-I} & \multirow{2}{*}{mR} \\
\cline{2-7}
\cline{9-14}
& r@1 & r@5 & r@10 & r@1 & r@5 & r@10& & r@1 & r@5 & r@10 & r@1 & r@5 & r@10& \\
\hline
\hline
\textbf{(a)} Prior& & & & & & & & & & & & & &\\
Trans. to En&44.8 &74.3& 85.4& 36.9& 71.0 &84.7 &66.1&-- & -- & -- & -- & -- & -- & -- \\
EmbN& 56.0 & 83.7 & 90.7 & 45.5 & 77.2 & 87.3 & 73.2 & -- & -- & -- & -- & -- & -- & --\\
PAR. EmbN& 60.1 & 86.0 & 92.8 & 47.7 & 79.6 & 89.7 & 76.0& -- & -- & -- & -- & -- & -- & -- \\
(3-4) MULE& 59.6 & \textbf{86.5} & 92.8 & 47.8 & 80.8 & 90.1 & 76.3 & -- & -- & -- & -- & -- & -- & --  \\
\hline
\hline
\textbf{(b)} Adapted& & & & & & & & & & & & & &\\
(1) S-LIWE&   57.2&85.0&93.2&42.2&76.4&87.6&73.6 
& -- & -- & -- & -- & -- & -- & --\\
(2) S-LIWE& 45.3&78.2&89.5&36.4&68.9&81.2&66.6
 & -- & -- & -- & -- & -- & -- & --\\
(10) S-LIWE& 45.9&77.9&88.2&34.1&67.5&81.2&65.8
 &   41.8 & 72.4 & 82.1 & 25.3 & 52.4 & 63.4 & 56.2
\\
(10) L-LIWE& 51.5&81.4&90.2& 39.1&71.4&84.3&69.6
&  40.5 & 71.0 & 82.1 & 26.2 & 53.5 & 64.7 & 56.3
 \\
(10) MULE& 59.4 & 85.2 & 93.0 & 47.4 & 80.1 & 90.2 & 75.9 & \textbf{49.9} & \textbf{80.2} & \textbf{87.7} & 38.1 &\textbf{69.3} & 78.6 & \textbf{67.3} \\
\hline
\hline
\textbf{(c)} Ours& & & & & & & & & & & & & &\\
LA & 51.4 & 83.3 & 90.3 & 42.4 & 76.8 & 88.1 & 72.1 & 48.4 & 77.2 & 85.7 & 35.5 & 65.1 & 76.5 & 64.7\\
HEM & 56.8 & 86.3 & 93.8 & 47.7 & 81.7 & \textbf{91.7} & 76.3& 48.9 & 78.4 & 86.0 & 38.4 & 68.0 & 78.3& 66.3\\
SMALR & 60.4 & 86.4& \textbf{94.3} & 48.5 & 82.2 & 91.2 & 77.2& 46.8& 79.1& 87.6 & \textbf{38.8} & 69.1 & \textbf{78.8} & 66.7\\
SMALR-A & 60.0 & 84.5& 92.9 & 45.9 &  78.3 & 88.6 & 75.0 & -- & -- & -- & -- & -- & -- & --\\
SMALR-C & \textbf{61.9} & 86.4 & 94.0 & \textbf{49.3} & \textbf{81.9} &  91.3 & \textbf{77.5}& -- & -- & -- & -- & -- & -- & -- \\
\hline
\end{tabular}
\end{center}
\end{table}

\begin{table}[ht]
\setlength{\tabcolsep}{1.pt}
\begin{center}
\caption{Arabic bidirectional image-sentence retrieval results using sentences translated from English into Arabic for testing}
\label{tab:mscoco_ar}
\begin{tabular}{|l|c|c|c|c|c|c|c|c|c|c|c|c|c|c|}
\hline
\multirow{3}{*}{Model} & \multicolumn{7}{c|}{MSCOCO} & \multicolumn{7}{c|}{Multi30K} \\
\cline{2-8}
\cline{9-15}
 & \multicolumn{3}{c|}{I-to-S} & \multicolumn{3}{c|}{S-to-I} & \multirow{2}{*}{mR} & \multicolumn{3}{c|}{I-to-S} & \multicolumn{3}{c|}{S-to-I} & \multirow{2}{*}{mR} \\
\cline{2-7}
\cline{9-14}
& r@1 & r@5 & r@10 & r@1 & r@5 & r@10& & r@1 & r@5 & r@10 & r@1 & r@5 & r@10& \\
\hline
\hline
\textbf{(a)} Adapted& & & & & & & & & & & & & &\\
(10) S-LIWE&  43.4&75.6&86.3&33.1&65.7&78.6& 63.8
 & 47.5 & 76.9 & 84.7 & 33.3 & 61.9 & 72.0 & 62.7 \\
(10) L-LIWE& 49.1&81.4&90.3&34.4&68.4&81.0& 67.5
 & \textbf{48.7} & 78.4 & \textbf{87.9} & 34.3 & 65.3 & 75.5 & 65.0
 \\
(10) MULE& \textbf{60.3} & 88.3 & 94.6 & 47.9 & 81.2 & 90.7 & 77.2&48.6& 78.2 & 87.4 & 36.7 & 66.8 & 76.9 & 65.8 \\
\hline
\hline
\textbf{(b)} Ours& & & & & & & & & & & & & &\\
LA & 56.1& 85.5& 93.6& 44.0& 78.3& 88.7& 74.4& 44.7 & 78.1 & 85.6& 34.5 & 65.2 & 75.3 & 63.9\\
HEM & 58.4 & 87.9 & 94.9 & 47.6 & 81.5 & 91.4 & 77.0& 45.9 & 76.8 & 85.6 & 36.3 & 66.2 & 76.2& 64.5\\
SMALR & 60.1 & \textbf{89.0} & \textbf{95.7} & \textbf{48.6} & \textbf{81.9} & \textbf{91.9} & \textbf{77.9} & 46.2 & \textbf{78.6} & 87.4 &\textbf{38.3} & \textbf{67.7} & \textbf{77.9} & \textbf{66.0} \\
\hline
\end{tabular}
\end{center}
\end{table}

\begin{table}[ht]
\setlength{\tabcolsep}{1.pt}
\begin{center}
\caption{Afrikaans bidirectional image-sentence retrieval results using sentences translated from English into Afrikaans for testing}
\label{tab:mscoco_af}
\begin{tabular}{|l|c|c|c|c|c|c|c|c|c|c|c|c|c|c|}
\hline
\multirow{3}{*}{Model} & \multicolumn{7}{c|}{MSCOCO} & \multicolumn{7}{c|}{Multi30K} \\
\cline{2-8}
\cline{9-15}
 & \multicolumn{3}{c|}{I-to-S} & \multicolumn{3}{c|}{S-to-I} & \multirow{2}{*}{mR} & \multicolumn{3}{c|}{I-to-S} & \multicolumn{3}{c|}{S-to-I} & \multirow{2}{*}{mR} \\
\cline{2-7}
\cline{9-14}
& r@1 & r@5 & r@10 & r@1 & r@5 & r@10& & r@1 & r@5 & r@10 & r@1 & r@5 & r@10& \\
\hline
\hline
\textbf{(a)} Adapted& & & & & & & & & & & & & &\\
(10) S-LIWE& 46.7 & 79.1
&  88.8
&  35.0
& 67.4
&  80.2
& 66.2
&  49.8 &  77.5 & 85.1 & 32.8 & 60.6 & 70.7 & 62.8\\
(10) L-LIWE& 49.9 & 82.2&91.8&36.9&70.6&82.4 & 68.9 & 49.5 & 77.4 & 86.3 & 34.0 & 62.3 & 72.6 & 63.7
 \\
(10) MULE& 62.4 & 88.1 & 94.8 & 48.7 & 81.5 & 91.0 &77.8
 &  51.3 & 80.2 & \textbf{87.7} & 39.0 & 67.7 & 77.7 & 67.3\\
\hline
\hline
\textbf{(b)} Ours& & & & & & & & & & & & & &\\
LA &55.2 & 85.1& 92.7& 45.7&79.9 &89.5 & 74.7& \textbf{51.5} & 78.9 & 86.5 & 37.8 & 67.2 & 77.2 & 66.5\\
HEM & 59.8 & 86.4 & 93.9 & 47.5 & 81.2 & 91.2 & 76.7& 47.6 & 79.3 & 87.4 & 38.7 & \textbf{69.2} & 78.9 & 66.8\\
SMALR & \textbf{62.5} &\textbf{88.7} & \textbf{95.9} & \textbf{48.8} & \textbf{82.2} & \textbf{91.4} & \textbf{78.2}&48.7& \textbf{79.7 }& 87.5 & \textbf{40.5} & 68.8 & \textbf{79.1} & \textbf{67.4}\\
\hline
\end{tabular}
\end{center}
\end{table}

\begin{table}[ht]
\setlength{\tabcolsep}{1.pt}
\begin{center}
\caption{Korean bidirectional image-sentence retrieval results using sentences translated from English into Korean for testing}
\label{tab:mscoco_ko}
\begin{tabular}{|l|c|c|c|c|c|c|c|c|c|c|c|c|c|c|}
\hline
\multirow{3}{*}{Model} & \multicolumn{7}{c|}{MSCOCO} & \multicolumn{7}{c|}{Multi30K} \\
\cline{2-8}
\cline{9-15}
 & \multicolumn{3}{c|}{I-to-S} & \multicolumn{3}{c|}{S-to-I} & \multirow{2}{*}{mR} & \multicolumn{3}{c|}{I-to-S} & \multicolumn{3}{c|}{S-to-I} & \multirow{2}{*}{mR} \\
\cline{2-7}
\cline{9-14}
& r@1 & r@5 & r@10 & r@1 & r@5 & r@10& & r@1 & r@5 & r@10 & r@1 & r@5 & r@10& \\
\hline
\hline
\textbf{(a)} Adapted& & & & & & & & & & & & & &\\
(10) S-LIWE& 42.4 & 76.4 & 86.9 & 31.9 & 63.6 &77.4 & 63.1& 38.7 & 69.1 & 80.2 & 24.3 & 51.8 & 62.7 & 54.5\\
(10) L-LIWE& 48.1& 79.8& 89.5 & 33.4& 66.3 & 79.9 & 66.2 & 41.2 & 73.5 & 83.2 & 26.3 & 53.4 & 65.0 & 57.1\\
(10) MULE& 56.5 & 85.6 & 93.5 & 43.9 & 78.0 & 88.5 &74.3&\textbf{47.1} & 76.1 & 85.3 & 35.0 & 63.7 & 74.6 & 63.6
  \\
\hline
\hline
\textbf{(b)} Ours&& & & & & & & & & & & & &\\
LA & 51.9&85.0 &92.2 &40.2 &73.8 &86.4 & 71.6& 43.3 & 73.4 & 83.0 & 31.3 & 59.4 & 71.1 & 60.3\\
HEM & \textbf{57.0} & 85.8 & 94.6 & \textbf{46.2} & \textbf{79.2} & \textbf{90.0} & \textbf{75.5}& 44.4 & 76.6 & 85.4 & 32.9 & 62.0 & 72.2& 62.3 \\
SMALR & 55.7 & \textbf{86.9} & \textbf{94.8} & 45.2 & 78.8 & 89.4 & 75.1 & 45.7 & \textbf{78.2} & \textbf{85.5} & \textbf{35.2} & \textbf{64.8} &\textbf{75.5}& \textbf{64.2}\\
\hline
\end{tabular}
\end{center}
\end{table}

\begin{table}[ht]
\setlength{\tabcolsep}{1.pt}
\begin{center}
\caption{Russian bidirectional image-sentence retrieval results using sentences translated from English into Russian for testing}
\label{tab:mscoco_ru}
\begin{tabular}{|l|c|c|c|c|c|c|c|c|c|c|c|c|c|c|}
\hline
\multirow{3}{*}{Model} & \multicolumn{7}{c|}{MSCOCO} & \multicolumn{7}{c|}{Multi30K} \\
\cline{2-8}
\cline{9-15}
 & \multicolumn{3}{c|}{I-to-S} & \multicolumn{3}{c|}{S-to-I} & \multirow{2}{*}{mR} & \multicolumn{3}{c|}{I-to-S} & \multicolumn{3}{c|}{S-to-I} & \multirow{2}{*}{mR} \\
\cline{2-7}
\cline{9-14}
& r@1 & r@5 & r@10 & r@1 & r@5 & r@10& & r@1 & r@5 & r@10 & r@1 & r@5 & r@10& \\
\hline
\hline
\textbf{(a)} Adapted& & & & & & & & & & & & & &\\
(10) S-LIWE& 44.7 & 76.1 & 86.5 & 31.4 & 64.5& 78.2& 63.6& 47.7 & 77.4 & 84.6 & 33.6 & 62.9 & 72.6 & 63.1 \\
(10) L-LIWE& 50.7& 83.6 & 91.5 & 36.7 & 71.7 & 83.0& 69.6& \textbf{49.7} & \textbf{79.5} & \textbf{86.9} & 35.2 & 65.2 & 76.0 & 65.4 \\
(10) MULE& 60.8 & \textbf{89.0} & 94.9 & 48.0 & 80.4 & 90.4 & 77.3 &  48.3 & 78.6 & 86.2 & 37.1& 65.8 & 76.2 & 65.4\\
\hline
\hline
\textbf{(b)} Ours&& & & & & & & & & & & & &\\
LA& 53.7 & 85.0 & 92.1 & 42.4 & 75.8 & 87.4 & 72.7& 42.2 & 72.7 & 82.8 & 31.6 & 60.7 & 71.7 & 60.3\\
HEM & 58.3 & 87.5 & 94.4 & \textbf{48.5} & \textbf{81.8} & \textbf{91.7} & 77.0&45.6 & 75.0 & 83.6 & 35.3 & 63.2 & 73.1 & 62.6\\
SMALR & \textbf{62.7} & 88.8 & \textbf{95.0} & 48.2 & 81.7 & 91.5 & \textbf{78.0}& 48.4 & 77.3 & 86.0 & \textbf{38.0} & \textbf{67.2} & \textbf{77.5} & \textbf{65.7}\\
\hline
\end{tabular}
\end{center}
\end{table}
\subsection{Testing with Machine Translations}
\label{subsec:trans_compare}
In this section we investigate the effect testing with machine translations rather than human-generated sentences has when comparing methods.  For all methods we use models trained on all 10 languages, and test on human-generated and translated sentences for Chinese and Japanese on MSCOCO and German, French, and Czech on Multi30k.

As seen below, there are only minor differences in the performance of each language we tested.  Notably, the performance rankings with each dataset are consistent regardless of whether the method is evaluated on human generated test sentences or test sentences translated from English.

\begin{table}[h]
\setlength{\tabcolsep}{1.pt}
\begin{center}
\caption{Comparison of using Human Generated Sentences vs. Translations for testing purposes.}
\label{tab:m30k_translations}
\begin{tabular}{|rl|c|c|c|c|c|c|c|c|c|}
\hline
& \multirow{2}{*}{Model}  & \multicolumn{4}{c|}{MSCOCO} & \multicolumn{5}{c|}{Multi30k}\\
\cline{3-11}
& &\multicolumn{2}{c|}{mR} & \multirow{2}{*}{Avg} & \multirow{2}{*}{Rank}  &\multicolumn{3}{c|}{mR} & \multirow{2}{*}{Avg} & \multirow{2}{*}{Rank}\\
\cline{3-4}
\cline{7-9}
& & Cn & Ja & & & De & Fr & Cs & &\\
\hline
\hline
\textbf{(a)} & Human generated test sentences & & & & & & & & &\\
& (10) S-LIWE~\cite{Wehrmann_2019_ICCV} & 64.5 & 65.8 & 65.2 & 6 & 65.2 & 51.8 & 50.0 & 55.7 & 6\\
& (10) L-LIWE & 70.0 & 69.6 & 69.8 & 5& \textbf{66.7} & 53.5 & 51.3 & 57.2 & 5 \\
& (10) MULE~\cite{kimMULEAAAI2020} & 75.6 & 75.9& 75.8& 3 & 63.6 & 63.4 & 59.4 & 62.1 &2\\
& Language-Agnostic & 72.3 & 72.1 & 72.2 & 4 & 61.3 & 59.9 & 54.0 & 58.4 & 4\\
& HEM & \textbf{76.7} & 76.3 & 76.5 & 2 & 62.8 & 63.3 & 60.0 & 62.0 & 3\\
& SMALR & \textbf{76.7}& \textbf{77.2}& \textbf{76.9} & 1 & 64.7 & \textbf{64.5} & \textbf{61.1} & \textbf{63.4} & 1\\
\hline\hline
\textbf{(b)} & Test sentences translated from En & & & & & & & & &\\
& (10) S-LIWE~\cite{Wehrmann_2019_ICCV} &  64.7 & 65.8 & 65.2 & 6 & 64.3 & 49.9& 52.1 & 55.4 & 6 \\
& (10) L-LIWE & 70.0 & 69.6 & 69.8 & 5 & \textbf{65.8} & 51.8 & 54.8 & 57.5 & 5\\
& (10) MULE~\cite{kimMULEAAAI2020} & 73.2& 75.0& 74.1 & 3 &64.4 & 64.0 & 64.8 & 64.4 & 2 \\
& Language-Agnostic & 69.7 & 71.4 & 70.6 & 4 & 61.7 & 61.1 & 60.5 & 61.1 & 4\\
& HEM & 73.5 & 75.3 & 74.4 & 2 & 63.8 & 63.5 & 64.3 & 63.9 & 3 \\
& SMALR & \textbf{74.4} & \textbf{75.9} & \textbf{75.2} & 1 & 65.1 & \textbf{65.1} & \textbf{65.5} & \textbf{65.2} & 1\\
\hline
\end{tabular}
\end{center}
\end{table}

\end{document}